\documentclass{article}

\usepackage{arxiv}

\usepackage[utf8]{inputenc} 
\usepackage[T1]{fontenc}    
\usepackage{hyperref}       
\usepackage{url}            
\usepackage{booktabs}       
\usepackage{amsfonts}       
\usepackage{nicefrac}       
\usepackage{microtype}      
\usepackage{lipsum}
\usepackage{graphicx}
\graphicspath{ {./images/} }
\usepackage{amsfonts}
\usepackage{amsmath}
\usepackage{amssymb}
\usepackage{dsfont}
\usepackage{graphicx}
\usepackage{subfigure}
\usepackage{subcaption}
\usepackage{textcomp}
\usepackage{color}
\usepackage{adjustbox} 
\usepackage{booktabs}
\usepackage{multirow}
\usepackage{soul}
\usepackage{comment}
\usepackage[table]{xcolor}
\usepackage{tikz}
\usetikzlibrary{matrix,positioning,arrows.meta}

\title{Auto-Configured Explainable Graph Neural Networks for Multi-Site Pollution Prediction}

\author{
Abdelkader Dairi \\
Department of Computer Science \\
University of Science and Technology of Oran-Mohamed Boudiaf (USTO-MB) \\
El Mnaouar, BP 1505, Bir El Djir 31000, Oran, Algeria \\
\texttt{abdelkader.dairi@univ-usto.dz}
\And
Fouzi Harrou \\
Computer, Electrical and Mathematical Sciences and Engineering (CEMSE) Division \\
King Abdullah University of Science and Technology (KAUST) \\
Thuwal 23955-6900, Saudi Arabia \\
\texttt{fouzi.harrou@kaust.edu.sa}
\And
Ying Sun \\
Computer, Electrical and Mathematical Sciences and Engineering (CEMSE) Division \\
King Abdullah University of Science and Technology (KAUST) \\
Thuwal 23955-6900, Saudi Arabia \\
\texttt{ying.sun@kaust.edu.sa}
}

\begin{document}
\maketitle
\begin{abstract}
Accurate particulate matter (PM) prediction is crucial for mitigating air pollution. Graph Neural Networks (GNNs) effectively model spatiotemporal dependencies, but predefined graphs limit adaptability, and some datasets complicate learning. This study introduces a graph construction method based on a confusion matrix from a supervised learning process to dynamically capture inter-class relationships. Additionally, a hybrid loss function that combines energy distance and Huber loss is applied to address the vanishing gradient problem and improve learning stability. The approach is evaluated using air pollution data from the University of Utah AirU Pollution Monitoring Network in Salt Lake City, UT, with five GNN models: Graph Convolutional Networks (GCNs), Simple Graph Convolutional Networks (SGConv), Graph Isomorphism Networks (GINs), Graph Attention Networks (GATs), and GraphSage. The experimental results of single- and multistep predictions confirm that GraphSage achieves the highest accuracy in predicting the concentrations of PM${1}$, PM${10}$, and PM$_{2.5}$ over different time horizons. Furthermore, {\color{black} GNNExplainer (Graph Neural Network Explainer) and PGExplainer (Probabilistic Graph Explainer)} are applied to interpret feature importance and graph structure, ensuring model transparency. Results show improved prediction accuracy, with GNN models outperforming traditional machine learning \textcolor{black}{and deep learning models (i.e., Prophet, Long short-term memory, Gated recurrent units} in air pollution forecasting.
\end{abstract}


\section{Introduction}\label{intro}
Air pollution, particularly particulate matter (PM), poses significant threats to public health and the environment~\cite{hernandez2020analysis}. PM, a complex mixture of tiny particles and liquid droplets, can penetrate deep into the respiratory system, causing various health problems such as asthma, bronchitis, heart attacks, and even premature death~\cite{polezer2018assessing}. Children, the elderly, and individuals with pre-existing health conditions are particularly vulnerable populations at risk~\cite{yang2019human}. Accurate prediction of PM concentrations is crucial for mitigating these health risks. Predictive models enable timely interventions and inform policy decisions to improve air quality, protect public health, and enhance environmental sustainability~\cite{gu2019stacked}. By forecasting PM levels, communities can better prepare to reduce exposure and take proactive measures, contributing to a healthier and safer living environment~\cite{dairi2021integrated}.

\medskip
Traditional methods for predicting PM concentrations have relied heavily on time series models and shallow machine learning methods. Time series models like seasonal decomposition of time series and autoregressive integrated moving average (ARIMA) are widely employed for their capability to capture temporal patterns in data~\cite{lee2012seasonal,harrou2013statistical}. However, these models often struggle with high-dimensional, non-linear relationships inherent in environmental data. They are typically limited by their reliance on historical data alone, failing to incorporate complex interactions between multiple factors that influence PM levels. Shallow machine learning methods, including linear regression, decision trees, and support vector machines (SVMs), have also been applied to PM prediction~\cite{wu2022machine}. These methods can capture linear and simple non-linear relationships, offering some improvements over traditional time series models. However, they generally fall short in handling the spatiotemporal dependencies and the intricate dynamics of air pollution. Shallow machine learning models lack the depth required to extract meaningful features from large, heterogeneous datasets, often leading to suboptimal performance in predicting PM concentrations.

\medskip
Graph Neural Networks (GNNs) have recently demonstrated significant potential in capturing complex relationships within environmental data~\cite{yang2022bearing,bloemheuvel2023graph}. GNNs can effectively capture the spatial and temporal dependencies inherent in air pollution data, offering a more nuanced and accurate approach to PM prediction. Various studies have demonstrated the superiority of GNN-based models over traditional methods.  Recent advances in spatio-temporal graph neural networks have reinforced their relevance in urban computing applications, including air pollution forecasting~\cite{jin2023spatio}. These models leverage both spatial correlations among monitoring stations and temporal dependencies in air quality data to enhance predictive performance. Furthermore, hybrid architectures integrating CNNs and adaptive GCNs have been proposed to address the limitations of purely distance-based approaches by capturing both geographic and latent region-wise dependencies~\cite{jin2022adaptive}. Additionally, automated spatio-temporal synchronous modeling frameworks have emerged to improve dynamic predictions, as demonstrated by Li et al.~\cite{li2022automated}, where multiple graph structures are leveraged to refine message-passing mechanisms for more robust traffic forecasting. For instance, Qi et al. introduced a hybrid model, GC-LSTM, which integrates Graph Convolutional Networks (GCN) with Long Short-Term Memory (LSTM) networks to forecast PM$_{2.5}$ concentrations~\cite{Qi_2019}. The model outperformed state-of-the-art methods by using spatiotemporal graph series from historical data and various air quality and meteorological factors as graph signals. It achieved a high correlation coefficient (R$^{2}$ = 0.72) for 72-hour predictions, demonstrating its potential for future pollutant concentration forecasting. In~\cite{Kim_2023},  Kim et al. developed a novel framework for PM$_{2.5}$ prediction using a multi-gated graph neural network. The model captured complex interactions between monitoring stations by employing multiple edges based on an atmospheric diffusion coefficient and PM$_{2.5}$ similarity metric. The model demonstrated significant improvements in root-mean-square error (RMSE) (2.613\%) and R$^{2}$ (5.263\%) for 96-hour predictions compared to conventional time-series models, highlighting its effectiveness in capturing global features and mitigating local dependency issues. In~\cite{Lin_2022}, Lin et al. introduced the ST-CCN-PM$_{2.5}$, a framework combining spatial attention mechanisms and causal convolution networks to enhance PM$_{2.5}$ prediction accuracy. The model outperformed several baseline models, showing a substantial decrease in RMSE (27.05\%), MAE (10.38\%), and an increase in R$^{2}$ (3.56\%) for single stations. The study in~\cite{Mandal_2023}
proposed the SA-GNN model for predicting short-term PM$_{2.5}$ concentrations by treating monitoring stations as graph nodes and leveraging their spatial relationships. The model incorporated meteorological variables and clustering-based spatiotemporal feature extraction within a graph neural network framework. Applied in Delhi, the model significantly improved R$^{2}$ (0.75), RMSE (25.13 $\mu g/m^{3}$), and MAE (21.28 $\mu g/m^{3}$), especially during high pollution episodes, demonstrating its potential for similarly polluted cities.

In~\cite{Ejurothu_2022}, Ejurothu et al. developed a Local Hybrid-Graph Neural Network (HGNN) approach for monitoring station-wise multi-step PM$_{2.5}$ forecasting across India. The model integrated spatiotemporal units and station-wise feature extraction units to handle local meteorological variations. In~\cite{Zeng_2022}, Zeng et al. introduced the STGODE-M model, employing tensor-based ordinary differential equations to capture spatial-temporal dynamics and build deeper networks. The model included air humidity as an auxiliary feature and used wind direction data for adjacency matrix construction. Evaluated on a dataset for home-based care parks, the STGODE-M showed superior performance in capturing spatial-temporal characteristics of PM$_{2.5}$, providing better guidance for elderly travel and reducing health risks. Liu et al.~\cite{liu2021new} presented a new benchmark task for graph-based machine learning, focusing on predicting future PM$_{2.5}$ concentrations under distribution shift. Their study revealed that GNN models suffer more from distribution shifts than non-graph-based models, emphasizing the need for special attention when deploying spatio-temporal GNNs in practice. In~\cite{Teng_2023}, Teng et al. developed the GNN\_LSTM model, which captured spatiotemporal correlations among monitoring sites to improve long-term PM$_{2.5}$ forecasting. The model showed significant performance improvements in the Beijing-Tianjin-Hebei region, particularly during polluted episodes. The inclusion of AOD features further enhanced prediction accuracy, emphasizing the importance of neighborhood site features for long-term air quality forecasting. In another study, Zhang et al.~\cite{Zhang_2023} proposed a novel method for long-term PM$_{2.5}$ prediction using a spatiotemporal graph attention recurrent neural network combined with a Grey Wolf optimization algorithm. The model effectively integrated spatial and temporal dependencies, providing improved prediction accuracy and robustness against variations in PM$_{2.5}$ concentrations. In~\cite{Pei_2022}, Pei et al. proposed the PMNet model, combining adaptive variational mode decomposition (AVMD) with a multivariate temporal graph neural network (MtemGNN). The model effectively extracted complex relationships between multivariate time series and demonstrated superior prediction performance compared to baseline models. Ablation experiments highlighted the significant contributions of AVMD, GRU, and MtemGNN to reducing MAE and improving prediction accuracy.

\medskip
This work introduces a novel approach for predicting PM concentration across multiple monitoring sites. The method leverages advanced graph analysis (improved Graph Neural Network) to capture the complex interplay between various factors (multivariate) that influence PM levels for several diameters, considering both their location (spatial) and how they change over time (temporal). This approach utilizes historical data (time series) from these sites to enhance prediction. The fundamental elements of our work can be outlined as follows:
\begin{itemize}
\item This study introduces a novel method for constructing graph structures using a confusion matrix derived from a supervised learning process. Traditional methods often rely on predefined graph structures that can limit a model's adaptability and performance due to their static nature. In contrast, our approach dynamically reflects the data's inherent relationships, capturing inter-class dependencies to provide a more adaptive and accurate representation of spatial and temporal dependencies. This dynamic graph construction adapts more effectively to changes and variations in the data, leading to improved prediction accuracy. By capturing subtle but crucial inter-class relationships, the model's predictive capabilities are significantly enhanced.
\item Furthermore,  the proposed approach employs a hybrid loss function that combines energy distance and Huber loss to address the vanishing gradient problem, a major challenge in training deep neural networks. By integrating these two loss functions, our method ensures more stable and efficient learning, resulting in better model performance. This hybrid approach balances robustness and sensitivity, crucial for handling outliers and providing smooth gradients during training, thereby enhancing the model's overall accuracy.
\item Moreover, the proposed approach is extensively evaluated using a real-world dataset and five different GNN models: GCN, GAT, GIN, SGConv, and GraphSage. Additionally, a comparison with traditional machine learning models, including k-Nearest Neighbors (kNN), Random Forest (RF), Extra Trees (ET), Decision Tree (DT), and Gradient Boosting (GB), has been conducted to assess the relative performance of GNNs in PM concentration prediction. \textcolor{black}{To further strengthen the evaluation, deep learning models such as Prophet, Long Short-Term Memory (LSTM), and Gated Recurrent Unit (GRU) have also been included to benchmark performance under similar forecasting conditions.} This evaluation includes both single and multi-step prediction experiments, thoroughly comparing model performance across various scenarios. Such a comprehensive evaluation provides robust validation of the approach, demonstrating its effectiveness.
\item Finally, we utilize two explainable AI techniques to interpret the model's decision-making process: GNNExplainer and PGExplainer. These tools thoroughly analyze feature importance and graph structure, offering insights into how the model arrives at its predictions. This enhances the transparency of the GNN models.
\end{itemize}

\medskip 
The subsequent sections of this paper are structured as follows. Section~\ref{sec2} briefly presents the basic concepts of the five GNN models considered in this study. Section~\ref{sec3} outlines the main steps of the proposed GNN strategy. Section~\ref{sec4} discusses the data used and presents the results of single and multi-step PM pollution predictions. Section~\ref{sec5} concludes the study and future lines of improvements.

\section{Preliminary Materials}\label{sec2}
\begin{figure}[h!]
\centering
\includegraphics[scale=0.8]{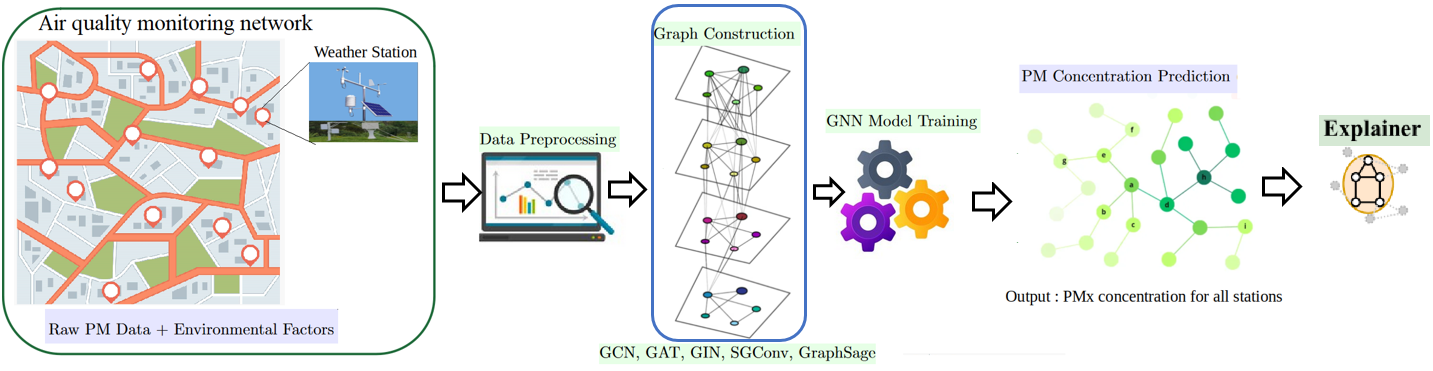} 
\caption{\color{black} Block diagram of the proposed GNN-Based PM prediction framework.}
\label{Framework}
\end{figure}

This section presents the materials used in the proposed approach, with a focus on GNNs and their variants. 
The block diagram in Figure~\ref{Framework} illustrates the proposed GNN-based framework for predicting particulate matter (PM) concentrations across multiple monitoring stations. The process begins with an air quality monitoring network, where raw PM data and environmental factors (e.g., temperature, humidity) are collected from weather stations distributed across an urban environment. In the data preprocessing stage, the collected data undergoes cleaning, normalization, and imputation to ensure consistency and quality. Next, a graph representation is constructed, where nodes represent monitoring stations and edges denote spatial or statistical relationships between them, determined using a confusion matrix-based approach. The constructed graph is then used to train multiple GNN models (GCN, GAT, GIN, SGConv, and GraphSage), which learn spatial-temporal dependencies in PM concentrations. Once trained, the PM concentration prediction module forecasts PM levels (PM$_{1}$, PM$_{2.5}$, PM$_{10}$) for all stations, using the graph structure to capture spatial correlations. Finally, to ensure model transparency, GNNExplainer and PGExplainer are applied for explainability analysis, identifying key features and relationships that contribute to the model’s predictions. This framework enables accurate, data-driven forecasting while ensuring interpretability through explainability techniques.

\subsection{Graph Neural Networks}
GNNs are a type of neural network specifically designed for processing graph-structured data. They excel at capturing the connections and dependencies among nodes within a graph, rendering them ideal for tasks including node classification, link prediction, and graph clustering. They leverage the connections between nodes (data points) to capture spatial relationships within the data. Unlike traditional neural networks, GNNs excel at processing graph-structured data by exploiting node connections to learn complex relationships. The basic concept of GNNs involves iteratively aggregating information from a node's neighbors to update its representation, allowing the network to learn from the structural information of the graph.

\medskip 
Five well-established GNN models are considered in this study: Graph Convolutional Networks (GCNs), Simple Graph Convolutional Networks (SGConv), Graph Isomorphism Networks (GINs), Graph Attention Networks (GATs), and GraphSage.

\medskip 
\paragraph{GCNs models}  pioneered the field of GNNs with their use of spectral convolutions to aggregate information from neighboring nodes~\cite{kipf2017semi}. GCNs are efficient and interpretable, but their reliance on spectral properties can limit their applicability to certain types of graphs. The GCN operation is defined as follows:
\begin{equation}
\mathbf{H(l+1)} = \sigma \left( \hat{\mathbf{D}}^{-\frac{1}{2}} \hat{\mathbf{A}} \hat{\mathbf{D}}^{-\frac{1}{2}} \mathbf{H(l)} \mathbf{W} \right),
\end{equation}
\noindent where $\hat{\mathbf{A}}$ is the adjacency matrix with self-connections added ($\hat{\mathbf{A}} = \mathbf{A} + \mathbf{I}$), $\mathbf{D}$ is the degree matrix of $\hat{\mathbf{A}}$, $\mathbf{H}$ is the input feature matrix, $\mathbf{W}$ is the learnable weight matrix, and $\sigma$ is a non-linear activation function (e.g., ReLU).

\paragraph{Simple Graph Convolutional Networks (SGConv)}
In the SimpleGConv layers of GNNs, a node's feature vector is enhanced by incorporating information from its neighbors~\cite{wu2019simplifying}. This is done by summing the feature vectors of neighboring nodes, each multiplied by a learned weight matrix. Additionally, the node's own feature vector undergoes a transformation using a separate weight matrix, and a bias vector is added for further customization. An activation function is applied next to introduce non-linearity, allowing the GNN to capture intricate relationships embedded within the graph structure. By integrating information from both the node itself and its neighbors, GNNs can learn rich representations of nodes that reflect the broader context of the entire graph.\\
 Simple GCNs do not involve message passing and aggregation in the usual GNN sense. Instead, they use a simplified approach where node features are directly averaged with features from neighboring nodes in the convolution operation. This averaging can be represented as:

\begin{equation}
H^{(l+1)}_i = \sigma \left(  \sum_{j \in \mathcal{N}_i} W^{(l)} H_j^{(l)} + WH_i^{(l)} + b^{(l)} \right),
\end{equation}

\noindent where $H^{(l)}_i$ is the feature vector of node $i$ at layer $l$, $W^{(l)}$ is the learnable weight matrix at layer $l$, $\sigma$ is a non-linear activation function (e.g., ReLU), $N(i)$ is the set of neighbor nodes of node $i$, and $|N(i)|$ is the degree (number of neighbors) of node $i$.

\paragraph{Graph Isomorphism Networks (GINs)}  are powerful for node classification tasks due to their message-passing framework and permutation equivariance~\cite{xu2018how}. They excel in distinguishing graph structures, although they might be less interpretable compared to GCNs. The GIN framework is represented by:
\begin{equation}
m_v^{(l+1)} = \varphi \left( \sigma \left( W^{(l)} \text{AGG} \left( { m_u^{(l)} \in M(u) | u \in N(v) }, h_v^{(l)} \right) \right) \right),
\end{equation}

\noindent where $N(v)$ is the set of neighbor nodes of node $v$, $W^{(l)}$ is a learnable weight matrix at layer $l$, $\sigma$ is a non-linear activation function (e.g., ReLU), and $\varphi$ is another learnable function that transforms the message after aggregation and activation.

\paragraph{GATs}  address the issue of uniform weighting in GCNs by introducing an attention mechanism, allowing the model to focus on informative neighbors~\cite{velickovic2017graph}. However, this can lead to higher computational costs. The attention mechanism in GATs can be represented as:
\begin{equation}
a_{vu}^{(l)} = \sigma(a(W_a^{(l)} h_v^{(l)}, W_a^{(l)} h_u^{(l)})),
\end{equation}
\noindent where $a_{vu}^{(l)}$ is the attention score for neighbor $u$ of node $v$ at layer $l$, and $a$ is an attention function that takes the features of node $v$ and its neighbor $u$ as input and outputs a raw attention score.

\medskip 
\paragraph{GraphSage}  explores inductive learning, making it suitable for large graphs and dynamic settings where unseen nodes might appear~\cite{wu2019comprehensive}. Defining an effective aggregation function remains a key consideration. The GraphSage aggregation process can be represented as:
\begin{equation}
m_v^{(l+1)} = AGG ( { m_u^{(l)} \in M(u) | u \in S_v^{(l)} }, h_v^{(l)}),
\end{equation}

\noindent where $AGG$ is a learnable function that combines messages and the node's own feature vector, $m_v^{(l+1)}$ is the aggregated message for node $v$ at layer $(l+1)$, $M(u)$ is the set of messages received by node $u$ at layer $l$, and $h_v^{(l)}$ is the feature vector of node $v$ at layer $l$. Common aggregation functions include sum, mean, and user-defined functions based on learnable neural networks.

\medskip
Overall, the five GNN models exhibit distinct strengths and trade-offs in capturing spatial and temporal dependencies for PM concentration forecasting. GCN effectively models local node relationships through spectral convolutions but may struggle with long-range dependencies. GAT improves upon this by incorporating an attention mechanism that assigns different importance weights to neighboring nodes, enhancing spatial feature extraction. GIN, designed to match the Weisfeiler-Lehman graph isomorphism test, provides strong graph representation learning but can be computationally demanding. SGConv simplifies graph convolution by reducing redundant transformations and improving computational efficiency while still effectively capturing both spatial and temporal dependencies through smooth information propagation across layers. This allows it to maintain structural coherence while preserving trends over time. GraphSage, with its inductive learning approach, excels in handling dynamic graphs by leveraging neighborhood sampling, making it highly efficient for real-world applications with evolving data.

\medskip
However, traditional GCNs rely on predefined graphs, which may not effectively capture dynamic relationships. While GATs adaptively weigh node importance and GINs enhance feature aggregation, these methods still depend on static graph structures. To address this limitation, an automatic graph construction approach based on a confusion matrix from supervised learning is introduced, enabling the model to better capture inter-class relationships. Additionally, a hybrid loss function is incorporated to enhance learning stability and mitigate vanishing gradient issues, ensuring more robust and accurate PM concentration forecasting.

\section{The proposed Methodology}\label{sec3}
\begin{figure}[h!]
\centering
\includegraphics[scale=0.5]{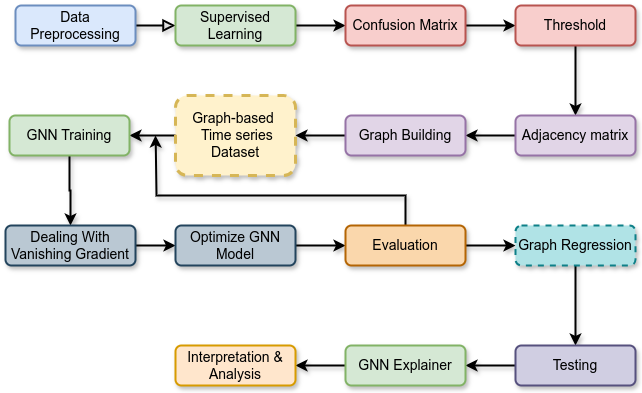} 
\caption{Schematic representation of the main steps in the proposed GNN-based approach.}
\label{Fig:flowchart}
\end{figure}

This section outlines the key steps of the proposed confusion matrix-based explainable GNN approach for multi-site pollution prediction, as illustrated in Figure (\ref{Fig:flowchart}). The methodology can be divided into several essential stages:
\begin{itemize}
\item Graph Construction: Utilizing a confusion matrix derived from supervised learning, we compute the adjacency matrix. This matrix captures the relationships between different pollution monitoring sites, forming the foundation of our graph structure.
\item GNN Training and Optimization: We train the GNN using the constructed graph. To address the vanishing gradient problem, we employ a hybrid loss function that combines the energy distance and Huber loss. This optimization step ensures robust learning and model stability.
\item Prediction: Once the GNN is trained, it is used for multi-site pollution prediction. The model leverages the spatial and temporal correlations captured during the graph construction and training phases to make accurate predictions of pollution levels across different sites.
\item Explainability and Interpretation: To interpret and analyze the predictions, we apply a GNN explainer. This tool helps to elucidate the model's decision-making process, providing insights into the factors influencing pollution levels and the interactions between different monitoring sites.
\end{itemize}
Figure~\ref{Fig:flowchart} illustrates the overall workflow of our approach, detailing each step from graph construction to model explainability.

\subsection{Data Preprocessing}
{\color{black}
Data preprocessing is a critical initial step to ensure the reliability, consistency, and accuracy of the predictive modeling pipeline. The process begins with data cleaning, which includes a thorough completeness analysis to remove inconsistent, noisy, or missing entries. Only monitoring stations with sufficient and continuous observations are retained to ensure that the input to the models reflects reliable spatial and temporal information. To handle missing values, the K-Nearest Neighbors (KNN) Imputer is employed. This imputation technique estimates missing data points by referencing the feature values of the $k$ most similar (nearest) observations based on Euclidean distance. This method preserves local data structure and correlation, making it suitable for environmental datasets with spatial dependencies. After cleaning and imputation, the dataset undergoes feature scaling to ensure uniformity in input ranges. All features are normalized using Min–Max normalization, which transforms the values of each feature into a common scale within the interval $[0,1]$. This step is crucial for improving the convergence behavior and stability of gradient-based optimization during neural network training. The Min–Max normalization is defined by:

\begin{equation}
x_i^{\text{scaled}} = \frac{x_i - x_i^{\min}}{x_i^{\max} - x_i^{\min}},
\end{equation}

\noindent
where $x_i$ represents the original value of the feature, and $x_i^{\min}$ and $x_i^{\max}$ are the minimum and maximum values of that feature in the training data, respectively. By applying this normalization technique, all variables contribute equally to the learning process, preventing features with larger numerical ranges from dominating model behavior. This standardization is particularly important when using models such as neural networks and GNNs, which are sensitive to scale differences across input features. }

\subsection{Graph Structure}
This study tackles the challenge of predicting PM concentration across multiple sites by employing an innovative graph-based method to analyze multivariate time series data. The process begins with constructing a graph to represent the relationships between monitoring stations. Traditionally, graphs are built using distance-based approaches, utilizing the GPS locations of monitoring stations. However, due to the absence of distance information in the dataset, an alternative method is necessary to establish connections within the graph.

To address this limitation, a novel approach leveraging supervised learning is introduced for creating the adjacency matrix. Specifically, a classification model predicts the device ID for each data point in the dataset using various features (pollutants) to distinguish between monitoring stations. This classification task produces a confusion matrix, where frequently confused classes are identified as candidates for graph connections. Essentially, classes with high misclassification rates are likely to be linked in the graph.

\subsubsection{Confusion matrix calculation}
The confusion matrix \(\mathbf{C} \in \mathbb{R}^{N \times N}\) is calculated based on a supervised classification task, where each data point is assigned a predicted class \( \hat{y} \) and compared to the true class \( y \). The matrix records how often class \( i \) is misclassified as class \( j \), the diagonal elements representing the correct classifications. Mathematically, the elements of the confusion matrix are defined as:
\begin{equation}
C_{i,j} = \sum_{k=1}^{M} \mathds{1} (y_k = i \land \hat{y}_k = j)
\end{equation}

\noindent
Where \( C_{i,j} \) represents the number of instances where class \( i \) was classified as class \( j \), \( M \) is the total number of data samples, \( \mathds{1} (\cdot) \) is the indicator function that returns 1 if the condition inside holds true and 0 otherwise, and \( y_k \) and \( \hat{y}_k \) are the true and predicted labels of the \( k \)-th data point.  The confusion matrix quantifies the misclassification patterns within the dataset, allowing the identification of inter-class relationships that inform graph construction.

\subsubsection{Constructing the Adjacency Matrix from the Confusion Matrix}
To transform the confusion matrix into an adjacency matrix \( \mathbf{A} \), a threshold \( \tau \) is applied to determine significant connections. A connection (edge) is established between two nodes if their confusion value exceeds a predefined empirical threshold:
\begin{equation}\label{CtoA}
    A _{i,j} = 
    \begin{cases}
        1, & \text{if } i \neq  j \text{ and } C_{i,j} > \tau, \\
        0, & \text{otherwise}.
    \end{cases}
\end{equation}

\noindent where \( A_{i,j} = 1 \) indicates the presence of an edge between node \( i \) and node \( j \), and \( \tau \) is a tunable parameter empirically determined to maintain connectivity while minimizing excessive noise.

This threshold ensures that only meaningful relationships based on frequent misclassification errors are used to form the graph structure while maintaining connectivity among all monitoring stations. As a result, the adjacency matrix serves as a blueprint to build a comprehensive and cohesive graph, encoding the connections between nodes (monitoring stations).

\subsection{Normalization of the Adjacency Matrix}
To stabilize the graph learning process, the adjacency matrix is normalized:

\begin{equation}
\tilde{\mathbf{A}} = \mathbf{D}^{-\frac{1}{2}} \mathbf{A} \mathbf{D}^{-\frac{1}{2}}.
\end{equation}

\noindent where \( \mathbf{D} \) is the degree matrix defined as \( D_{ii} = \sum_j A_{ij} \), and \( \tilde{\mathbf{A}} \) is the normalized adjacency matrix used in the GNN calculations. These normalized versions, represented by \( \tilde{\mathbf{A}} \), improve numerical stability and facilitate better information propagation across the graph, improving the efficiency of GNN training.

\medskip 

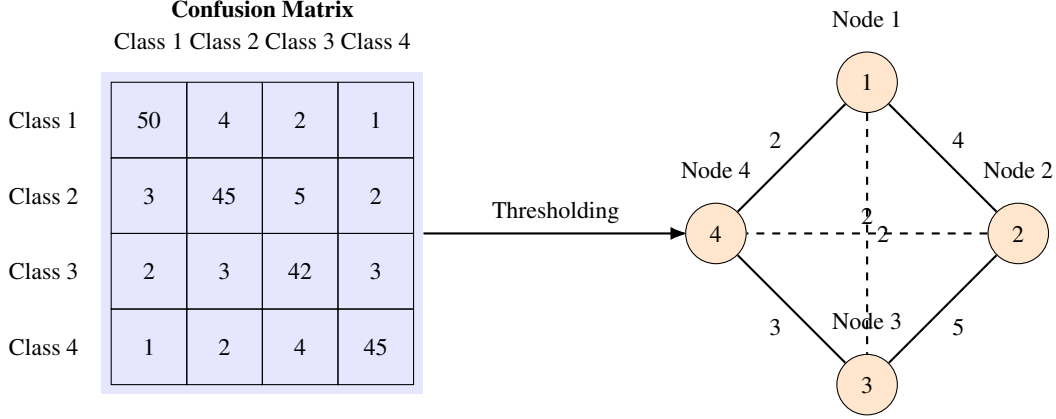
\begin{figure}[h!]
    \centering
    \begin{tikzpicture}[node distance=2cm, font=\small]

        \matrix (cmat) [matrix of nodes, nodes={draw, minimum size=1cm, anchor=center},
                        column sep=-\pgflinewidth, row sep=-\pgflinewidth, fill=blue!10] {
            50 & 4 & 2 & 1\\
            3 & 45 & 5 & 2\\
            2 & 3 & 42 & 3\\
            1 & 2 & 4 & 45\\
        };
        \node[above=0.6cm of cmat] {\textbf{Confusion Matrix}};

        \foreach \i in {1,...,4} {
            \node[left=0.3cm of cmat-\i-1] {Class~\i};
            \node[above=0.3cm of cmat-1-\i] {Class~\i};
        }

        \draw[thick,->,>=Latex] (cmat.east) -- ++(3.5cm,0) node[midway,above]{Thresholding};

        \begin{scope}[xshift=8cm, nodes={circle, draw, fill=orange!20, minimum size=0.8cm}]
            \node (N1) at (90:2) {1};
            \node (N2) at (0:2) {2};
            \node (N3) at (270:2) {3};
            \node (N4) at (180:2) {4};
        \end{scope}

        \draw[thick] (N1)--(N2) node[midway,above right] {4};
        \draw[thick,dashed] (N1)--(N3) node[midway,right] {2};
        \draw[thick] (N2)--(N3) node[midway,below right] {5};
        \draw[thick] (N3)--(N4) node[midway,below left] {3};
        \draw[thick] (N4)--(N1) node[midway,above left] {2};
        \draw[thick,dashed] (N2)--(N4) node[midway,above] {2};

        \foreach \i in {1,...,4} {
            \node[above=0.2cm of N\i] {Node \i};
        }

    \end{tikzpicture}
    \caption{ Illustrating the transformation from a confusion matrix (left) into a graph structure (right). Solid edges indicate strong relationships based on misclassifications, while dashed edges represent weaker yet relevant connections.}
    \label{fig:confusion-to-graph}
\end{figure}

Figure~\ref{fig:confusion-to-graph} illustrates the process of constructing a graph from a confusion matrix derived from a supervised classification task. The confusion matrix (left) represents the misclassification frequencies among four classes (Class 1 to Class 4). Each entry \( C_{i,j} \) in this matrix indicates the number of times samples from Class \( i \) were misclassified as Class \( j \). Higher misclassification values suggest stronger similarity or dependency between the corresponding classes.

To construct the graph (right), a thresholding operation is applied to filter significant misclassification frequencies, ensuring that only meaningful relationships contribute to graph formation. Nodes in the graph represent different classes, and edges between them reflect the misclassification relationships. The weight of an edge corresponds to the frequency of misclassification between two classes. Stronger connections (solid edges) represent higher misclassification frequencies, while weaker connections (dashed edges) indicate lower but still relevant relationships.

\medskip
This study focuses on establishing a robust graph structure based on the confusion matrix-derived adjacency matrix and does not consider edge features that could capture specific details about the relationships. The primary aim is to represent the interactions and dependencies among the monitoring stations effectively, facilitating accurate PM concentration predictions.

\subsection{Graph-based regression}
Graph-based regression utilizes the structure and features of a graph to predict continuous values for nodes or the entire graph \cite{ruiz2020gated,yao2024gstgat,HUANG2023261}. Unlike graph classification, which assigns discrete labels to nodes or graphs, graph regression predicts continuous outcomes \cite{zhang2023temporal,yin2023multi}. This method is akin to traditional multi-step forecasting, which considers interwoven relationships and influences between multiple variables over time.

\medskip 
This study addresses the challenge of predicting PM concentration levels (PM${1}$, PM${2.5}$, and PM${10}$) collected from several air quality monitoring stations. The approach utilizes six key features: temperature, humidity, MicsRED, MicsNOX, MicsHeater, and historical data on the target pollutant. This enables the Graph Neural Network (GNN) to capture how air pollution travels and influences surrounding areas. By iteratively exchanging information across the network (message passing), each station's representation is enriched, accounting for the influence of its upwind and downwind neighbors. A traditional regression layer then uses this enriched representation to predict PM concentrations (PM${2.5}$, PM${1}$ and PM${10}$) at each station. This approach allows GNNs to outperform traditional methods by considering the crucial spatial relationships between monitoring stations in PM forecasting.

\medskip 
To address challenges such as the vanishing or exploding gradient problem due to potentially long paths of information travel through the graph, a batched-graph learning approach is employed for model training, leveraging the graph structure. The performance of five different GNN models—SimpleGCN, GCN, GIN, GAT, and GraphSage—is evaluated. During training, the vanishing gradient problem emerged as a recurring challenge, impacting model convergence. To mitigate this, a hybrid loss function $\mathcal{L}(y, \hat{y})$ is designed, combining Huber loss $\mathcal{H}$ and Energy distance $\mathcal{E}$:
\begin{equation}\label{eq:huber}
    \mathcal{H}_{\delta}(y, \Hat{y}) = \left\{
  \begin{array}{ll}
    \frac{1}{2} (y - \Hat{y})^2 & \text{for} |y - \Hat{y}| \leq \delta \\
    \delta |y - \Hat{y}| - \frac{1}{2} \delta^2 & \text{for} |y - \Hat{y}| > \delta
  \end{array}
\right.
\end{equation}

\begin{equation}\label{eq:energy}
\mathcal{E}(u,v) = \Bigg( 2\int_{-\infty}^{\infty} \left( F_u(x) - F_v(x) \right)^p \Bigg)^{1/p}
\end{equation}
\begin{equation}\label{eq:loss}
 \mathcal{L}(y, \Hat{y}) = \mathcal{E}(y, \Hat{y}) * \mathcal{H}_{\delta}(y, \Hat{y})
\end{equation}

\noindent Here, $\mathcal{H}$ represents the Huber loss, $y$ denotes the true value, $\Hat{y}$ the predicted value, and $\delta$ is a parameter controlling the transition between quadratic and linear parts of the loss function. $\mathcal{E}$ is the energy distance between two distributions, with $p$ influencing the sensitivity to outliers.

\medskip 
Message passing is a fundamental concept in GNNs, enabling nodes to learn from the graph's structure by aggregating information from neighboring nodes. Through multiple message-passing steps across layers, GNNs progressively build richer representations of each node, taking into account both the node's features and the context provided by its connected neighbors. This approach can be framed as a time series regression problem, where the model learns from past observations in the station's data to predict future values. The model effectively captures temporal and spatial dependencies by processing data at the node level and incorporating other relevant features.

\medskip
The prediction performance of the five GNN architectures was compared on real-world datasets to assess the effectiveness of the PM spatiotemporal forecasting approach. The effectiveness of each prediction model was evaluated using statistical measures: RMSE, Mean Absolute Error (MAE), and $R^2$. 

\subsection{Graph regression Explainability}
While GNNs excel at graph regression tasks, their "black-box" nature can be problematic due to their complex message passing and non-linear activation functions, which make it difficult to directly interpret their reasoning. Despite their strong performance and wide applicability, GNN models lack transparency, making it challenging to understand the reasoning behind their outputs.

\medskip
Instance-level explanations are the dominant technique used in explainability \cite{yuan2022explainability,longa2022explaining,agarwal2023evaluating} analysis to understand how models arrive at specific predictions. These methods provide input-specific explanations tailored to each graph, identifying the key input features that drive the model's predictions. By focusing on input features, these methods generate input-dependent explanations that pinpoint the features most impactful on the model's output for each graph.

\medskip
This study utilizes the GNNExplainer \cite{ying2019gnnexplainer}, a post-hoc explanation method that focuses on individual instances (instance-based) and falls under the category of perturbation-based techniques. GNNExplainer helps elucidate the inner workings of GNNs by identifying the most relevant input features and connections that contribute to the model's predictions, thereby enhancing the interpretability of graph regression models.\\
A Parameterized Explainer for GNNs \cite{luo2020parameterized} is a learnable model designed to interpret GNN predictions by identifying key parts of the input graph that influence the model's decisions. It involves training an additional neural network that takes the original graph and the trained GNN as inputs, outputting importance scores for nodes and edges. These scores highlight the most influential substructures within the graph. The training process ensures the explainer aligns with the GNN's predictions, balancing sparsity and fidelity. This method enhances the transparency and trustworthiness of GNNs, making them more interpretable and suitable for critical applications.

\section{Results and discussion}\label{sec4}
\subsection{Data description}
The used dataset in this study contains air pollution measurements from 25 monitors in Salt Lake City, Utah, collected by the AirU Pollution Monitoring Network from January 1, 2019, to May 19, 2021~\cite{aeh2-a413-22}. Each pollution monitor transmits data packets every minute and is equipped with a suite of environmental sensors: a Plantower PMS3003 for counting airborne particles, a Texas Instruments HDC1080 for measuring temperature and humidity, and an SGX SensorTech MiCS4514 for detecting oxidizing and reducing gases. The dataset includes basic weather data, such as temperature and humidity, alongside readings from specialized sensors (MicsRED and MicsNOX) capable of detecting various gases, including CO, H2S, Ethanol, Ammonia, Hydrogen, Methane, and Propane. There is also a column indicating the state of a heater for the oxidizing sensor (MicsHeater). Most importantly, the dataset contains measurements of particulate matter (PM) in different sizes (PM$_{1}$, PM$_{2.5}$, and PM$_{10}$). Each pollution monitor is identified by a unique Device ID.

\subsection{Experiments and Settings}
To construct the graph, the adjacency matrices must first be computed. This is achieved by applying an XGBoost classifier~\cite{chen2016xgboost} to predict the device ID for each data point in the dataset, using various features (pollutants) to distinguish between monitoring stations. 

\begin{figure}[h!]
\centering
\includegraphics[width=11cm]{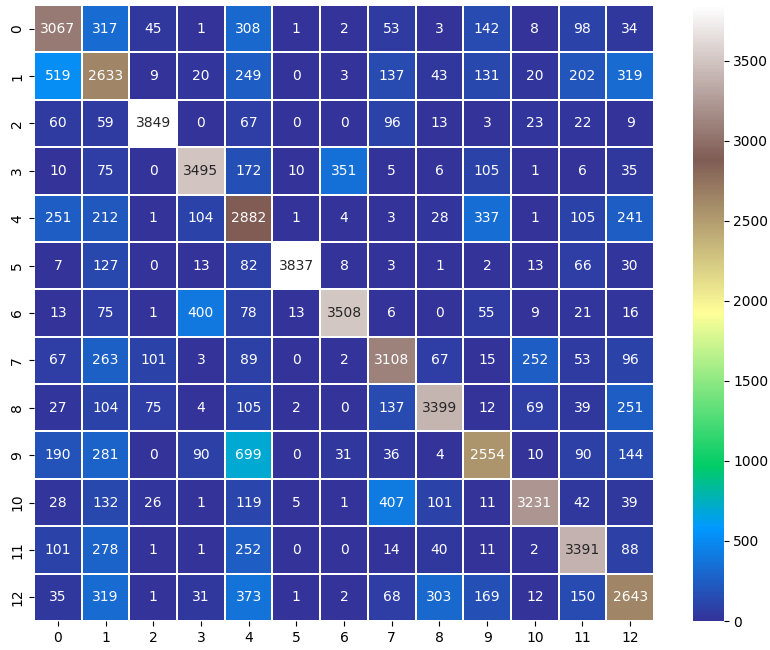} 
\caption{Confusion Matrix Result from Supervised Learning Using an XGBoost Classifier, Where Data Points in the Dataset Are Mapped to Device ID the target classes.}
\label{fig:confusion}
\end{figure}

For supervised learning with the XGBoost classifier, 80\% of the data was used for training and validation, while the remaining 20\% was reserved for testing. The model was configured with 500 trees, a learning rate of 0.1, 13 classes, and a maximum depth of 4. The "multi:softprob" objective was selected to output a probability distribution across classes, ensuring accurate classification of data points to their respective device IDs. This mapping is essential for constructing a robust adjacency matrix for the GNN. The classification task produces a confusion matrix (Figure~\ref{fig:confusion}), where frequently misclassified classes indicate strong inter-class relationships. These relationships inform graph construction, ensuring that nodes (monitoring stations) are linked based on data-driven spatial correlations, rather than arbitrary predefined connections.

\medskip 
The inclusion of data from multiple locations plays a crucial role in building the GNN by enabling the model to effectively learn spatial dependencies across diverse environments. As shown in Figure~\ref{fig:confusion}, the confusion matrix generated from the XGBoost classifier guides the graph construction process, ensuring that the resulting graph structure is data-driven rather than predefined. This approach captures meaningful relationships between different monitoring stations, allowing the model to dynamically reflect real-world spatial correlations.

\begin{figure}[h!]
\centering
\includegraphics[width=8cm]{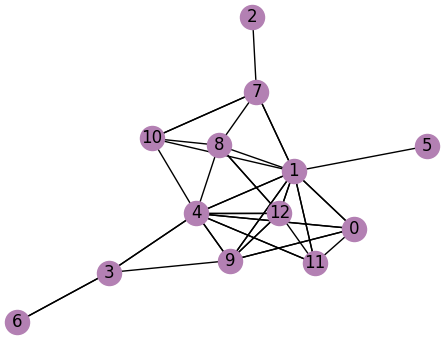} 
\caption{The graph constructed based on a thresholded confusion matrix.}
\label{fig:graph}
\end{figure}

\medskip 
Next, adjacency matrices are constructed using thresholded confusion matrices (refer to Equation~\ref{CtoA}). This approach takes advantage of the inherent relationships within the confusion matrix to encode the graph structure. The threshold ensures that all monitoring stations in the network remain connected, providing a comprehensive and cohesive graph structure (Figure~\ref{fig:graph}). The resulting adjacency matrix serves as a blueprint for constructing the graph, encoding the connections between nodes (monitoring stations).  The empirical threshold $\tau$ is selected by gradually increasing the value from the minimum non-zero entry of the confusion matrix until the resulting graph $G$ remains fully connected. This ensures that all classes are represented in the graph without excessive noise from low-frequency confusion. The value of $\tau$ is thus the smallest threshold that maintains graph connectivity. This data-driven selection process avoids arbitrary cutoff values and adapts to the structure of the supervised confusion matrix. In cases where no single threshold maintains connectivity, the fallback is a minimal spanning structure based on the strongest connections. This strategy balances interpretability, sparsity, and robustness in graph construction.

\begin{equation}
\tau = \begin{cases}
\min \{ i \in \{0, \dots, N-1\} \mid \text{connected}(G) \} & \text{if } \exists i \text{ where } \text{connected}(G), \\
\text{-1} & \text{otherwise.}
\end{cases}
\end{equation}

\medskip
After constructing the GNN graph based on the computed confusion matrix, the prediction performance of five GNN models (i.e., SimpleGCN, GCN, GIN, GAT, and GraphSage) in forecasting PM levels was assessed. Two series of experiments were conducted for this evaluation.
The first series focused on one-hour-ahead multivariate forecasting using the GNN models. The second series evaluated univariate forecasting of PM concentrations for short to medium-term horizons, specifically 3, 6, 9, and 12 hours ahead. For all GNN models employed in this study, the dataset was split into training and testing sets. In particular, 85\% of the data was dedicated to training and validation, while the remaining 15\% was allocated for testing. The training period spanned from January 1, 2019, to January 9, 2021, and the testing period ran from January 10, 2021, to May 19, 2021. This division ensured a robust evaluation of the models' predictive performance over different time horizons and spatiotemporal configurations.

\medskip
The GNNs investigated in this study were implemented using the Deep Graph Library (DGL)~\cite{wang2019deep}, a Python framework designed for efficient and flexible development of graph-based models. Based on PyTorch, DGL provides fine-grained control over message passing operations and supports performance optimizations such as auto-batching and sparse matrix kernels. These features allow for scalable and efficient training across multiple CPU and GPU environments. DGL’s modular design facilitated the implementation of all five GNN architectures considered in this work, including GCN, GAT, GIN, SGConv, and GraphSage. In addition, the NetworkX library \cite{hagberg2008exploring} was used alongside DGL to construct and verify graph connectivity, leveraging its utilities for graph structure analysis and preprocessing.  The hyperparameters for the GNN models adopted in the study were determined through a grid search approach. Grid search is a foundational technique for hyperparameter optimization in deep learning, facilitating a structured exploration of the hyperparameter space and leading to the identification of well-performing parameter configurations. The hyperparameters identified through this method are detailed in Table~\ref{tab:gnn-params}. Each model configuration includes layers specific to the GNN type followed by a fully connected layer, and common hyperparameters for optimization and training.
\begin{table}[h!]
    \centering
    \caption{Hyperparameters used for the GNN models in the study.} \label{tab:gnn-params}
    \begin{tabular}{c | l}
    \toprule
        Model & Parameters \\\toprule
         GCN & 03 Layers GCN, 01 layer: Fully Connected \\
         SGConv & 02 Layers SGConv, 01 layer: Fully Connected\\
         GIN & 02 Layers GIN, 01 layer: Fully Connected\\
         GSage & 02 Layers GSage, 01 layer: Fully Connected\\
         GAT & 02 Layers GAT, 01 layer: Fully Connected \\\toprule
                & Optimizer: Adam, epochs:500\\ 
         Common & hidden units:24, batch size: 512 \\ 
                &  activation function: ReLU \\
         \toprule
    \end{tabular}
        \end{table}

\begin{figure}[h!]
    \centering
\includegraphics[width=0.475\linewidth]{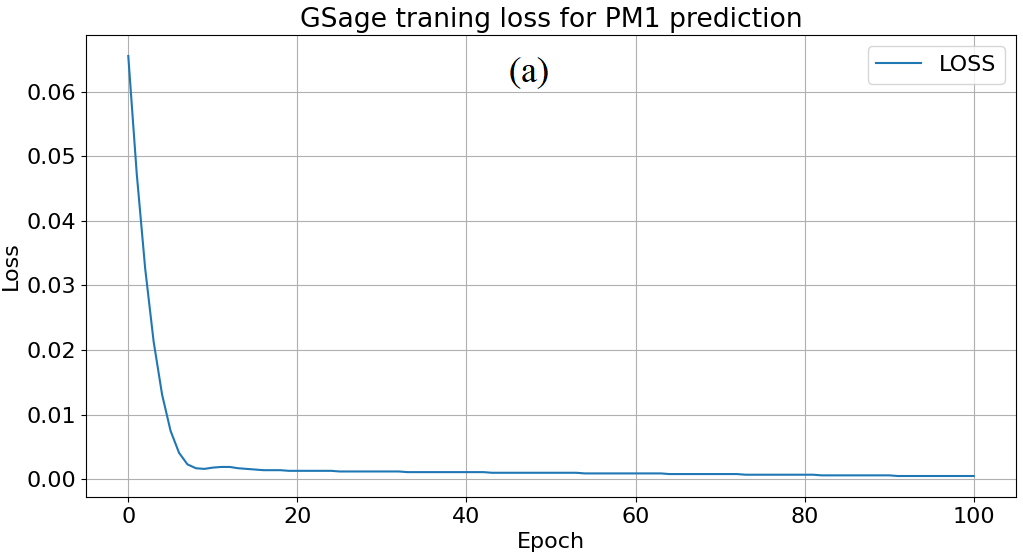}
\includegraphics[width=0.475\linewidth]{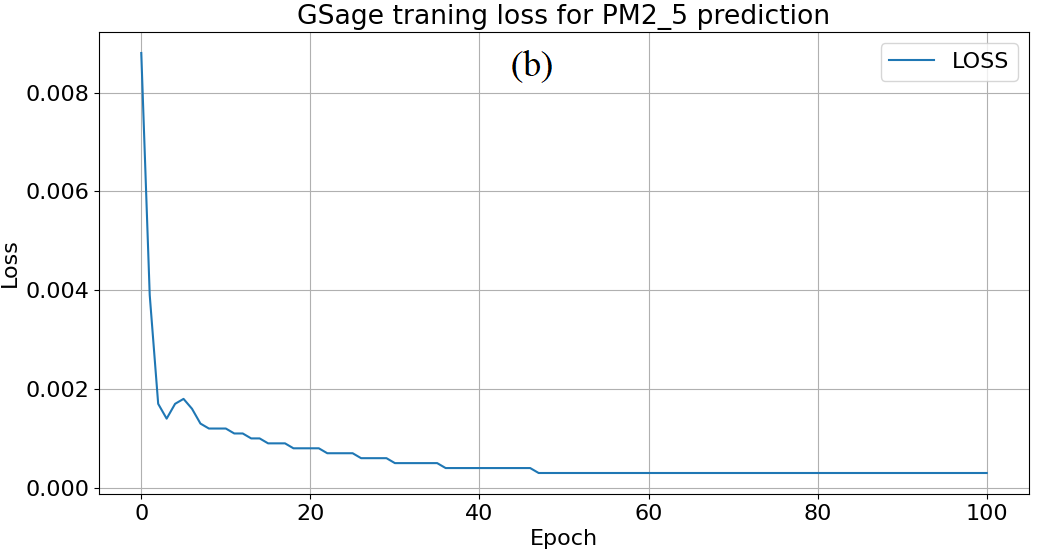}
\includegraphics[width=0.475\linewidth]{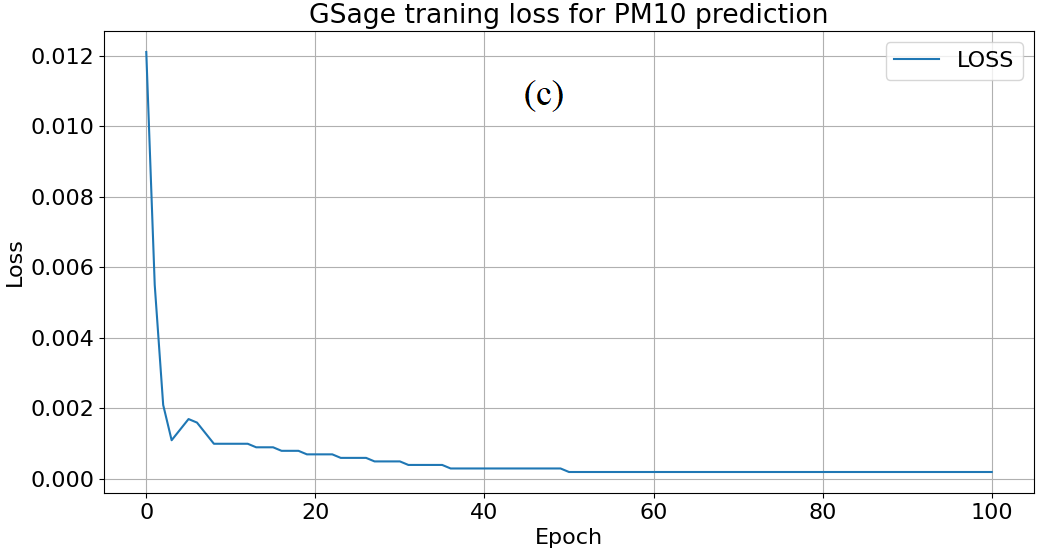}
         \caption{ Loss function curves for GSage training across different pollutant concentration predictions. (a) Training loss for PM$_{1}$ prediction, (b) Training loss for PM$_{2.5}$ prediction, and (c) Training loss for PM$_{10}$ prediction. The loss function exhibits smooth convergence, ensuring stable learning for all pollutant concentrations.}
    \label{fig:PM-GSage-loss}
    \end{figure}

During training, hyperparameters were selected to ensure stable loss function convergence, preventing overfitting and optimizing predictive accuracy. The tuning process involved iterative adjustments until the model demonstrated consistent performance across multiple training runs. The primary criterion was the smooth and rapid convergence of the loss function, indicating effective learning. For illustration, Figure~\ref{fig:PM-GSage-loss} (a-c) presents the training loss curves for the GraphSage model across different pollutant concentration predictions. The curves show that the model achieved stable convergence within a few epochs, confirming the effectiveness of the selected hyperparameters in preventing issues such as exploding or vanishing gradients.

\medskip
This study compares the performance of GNN-based models with traditional machine learning (ML) approaches for PM pollutant concentration forecasting. To ensure a rigorous evaluation, five widely used ML regression models, including k-Nearest Neighbors (kNN), Random Forest (RF), Extra Trees (ET), Decision Tree (DT), and Gradient Boosting (GB), were considered alongside the investigated GNN models. These ML models were selected based on their effectiveness in capturing nonlinear trends in time series forecasting applications. The hyperparameters for the ML models (Table~\ref{tab:ml-hyperparams}) were optimized to enhance predictive accuracy. Aligning training and evaluation strategies ensures a comprehensive assessment of their effectiveness in capturing air pollution patterns. Additionally, to provide a broader benchmark, deep learning (DL) models, Prophet, Long Short-Term Memory (LSTM), and Gated Recurrent Unit (GRU), were also included. These models are widely used for univariate and multivariate time series forecasting due to their ability to capture temporal dependencies. Their inclusion allows for a more complete comparison against both GNN and traditional ML models. For the deep learning baselines, the LSTM and GRU models were implemented with two hidden layers, each containing 64 units. The ReLU activation function was used, with a batch size of 64 and the mean squared error (MSE) as the loss function. The Adam optimizer was employed to train the models efficiently. For the Prophet model, automatic configuration was used for yearly, weekly, and daily seasonality, with a changepoint prior scale set to 0.01 to ensure conservative trend estimation. The seasonality mode was set to multiplicative, making the model suitable for short-term pollutant concentration forecasting. All experiments in this study were conducted on a personal laptop equipped with an Intel Core i7 8th Generation CPU, 16 GB of RAM, and an NVIDIA GeForce GTX 1050 GPU with 4 GB of VRAM. This configuration was sufficient to train and evaluate all machine learning, deep learning, and graph-based models used in this research without requiring access to high-performance computing infrastructure.

\begin{table}[h!]
\centering
\caption{ Hyperparameter configurations for the machine learning models used in PM concentration forecasting.}
\label{tab:ml-hyperparams}
\begin{tabular}{l|l}
\toprule
\textbf{Model} & \textbf{Hyperparameters} \\
\midrule
kNN & n\_neighbors=5, weights='uniform', algorithm='ball\_tree', leaf\_size=30, p=2, metric='minkowski' \\
RF & n\_estimators=100, criterion='squared\_error', splitter='best' \\
ET & n\_estimators=100, criterion='squared\_error', splitter='best' \\
DT & criterion='squared\_error', splitter='best' \\
GB & loss='squared\_error', learning\_rate=0.1, n\_estimators=100, criterion='friedman\_mse', \\
   & max\_depth=3, alpha=0.9, validation\_fraction=0.1, tol=0.0001 \\
\bottomrule
\end{tabular}
\end{table}

\subsection{Prediction results}
This section evaluates the predictive performance of GNN, ML, and DL models (Prophet, LSTM, GRU) for PM concentration forecasting. The assessment covers both single and multi-step predictions, analyzing how well each model captures temporal and spatial dependencies in air pollution data. It is structured into two main parts: the first examines the models' accuracy in forecasting pollutant concentrations for the next immediate time step, while the second extends the evaluation to longer forecasting horizons (3, 6, 9, and 12 hours ahead). The comparative analysis provides a comprehensive evaluation of each approach’s effectiveness and limitations in modeling air pollution dynamics.

\subsubsection{Single-Step PM prediction using GNN models}
The trained GNN models were evaluated on the testing dataset to predict PM${1}$, PM${2.5}$, and PM$_{10}$ concentrations. Table~\ref{TabSingle} presents the results of this hourly-based forecasting, comparing the performance of different GNN models using three statistical metrics: $R^{2}$, MAE, and RMSE.

\medskip 
The results in Table~\ref{TabSingle} demonstrate the predictive capabilities of five GNN models: GAT, GCN, SGConv, GSage, and GIN in forecasting PM${1}$, PM${2.5}$, and PM$_{10}$ concentrations. GSage consistently outperformed other models across all pollutants, achieving the highest $R^{2}$ values and the lowest MAE and RMSE values.

For PM${1}$ predictions, GSage recorded an $R^{2}$ of 0.9980, along with the lowest MAE (0.1641) and RMSE (0.2120), indicating superior accuracy. SGConv followed with an $R^{2}$ of 0.9901, while GIN, GAT, and GCN showed relatively lower performance. Similarly, for PM${10}$ predictions, GSage again achieved the highest accuracy with an $R^{2}$ of 0.9970, MAE of 0.3346, and RMSE of 0.4527. SGConv and GCN performed well, while GIN exhibited the highest errors. For PM$_{2.5}$, GSage maintained its top performance with an $R^{2}$ of 0.9973, MAE of 0.3065, and RMSE of 0.3780. SGConv and GAT produced competitive results, but GIN had the highest MAE and RMSE, indicating a lower predictive capacity.
\begin{table}[h!]
\centering
\caption{Performance comparison of GNN models for pollutant concentration prediction (PM$_{1}$, PM$_{2.5}$, and PM$_{10}$).} \label{TabSingle}
\begin{tabular}{ll|ccc}
\toprule
Pollutant & Model & R2 & MAE & RMSE \\\toprule
 & GAT & 0.9844 & 0.3734 & 0.5950 \\
 & GCN & 0.9838 & 0.4407 & 0.6051 \\
PM$_{1}$ & SGConv & 0.9901 & 0.3382 & 0.4741 \\
 & GSage & 0.9980 & 0.1641 & 0.2120 \\
 & GIN & 0.9874 & 0.3419 & 0.5347 \\\toprule
 & GAT & 0.9819 & 0.6025 & 1.1066 \\
 & GCN & 0.9840 & 0.6481 & 1.0419 \\
PM$_{10}$ & SGConv & 0.9890 & 0.6729 & 0.8622 \\
 & GSage & 0.9970 & 0.3346 & 0.4527 \\
 & GIN & 0.9810 & 0.8850 & 1.1344 \\\toprule
 & GAT & 0.9823 & 0.5401 & 0.9665 \\
 & GCN & 0.9801 & 0.7191 & 1.0249 \\
PM$_{2.5}$ & SGConv & 0.9912 & 0.4422 & 0.6816 \\
 & GSage & 0.9973 & 0.3065 & 0.3780 \\
 & GIN & 0.9778 & 0.9414 & 1.0837 \\\toprule
\end{tabular}
\end{table}

{\color{black}
Table~\ref{tab:ml-dl-performance} presents the predictive performance of classical machine learning (ML), time-series (Prophet), and deep learning (DL) models (LSTM and GRU) across PM${1}$, PM${2.5}$, and PM${10}$. Ensemble-based ML models (GB, ET, RF) consistently outperformed simpler models like DT and kNN across all pollutants, with GB and ET achieving the highest R$^2$ values (up to 0.892 for PM${2.5}$ and PM${10}$, and 0.886 for PM${1}$), demonstrating strong generalization capabilities. Prophet, while designed for time-series forecasting, showed lower accuracy compared to the top-performing ML models, particularly for PM${2.5}$ (R$^2$ = 0.669), indicating its limited suitability for this multivariate setup. Similarly, deep learning models such as LSTM and GRU underperformed relative to GB and ET, with LSTM reaching R$^2$ values of 0.680 for PM${2.5}$ and 0.689 for PM$_{1}$, while GRU exhibited slightly lower performance. These results suggest that although DL models are effective in capturing temporal dynamics, they may require larger datasets or architectural tuning to match ensemble ML accuracy in this context. Compared to the GNN-based results in Table~\ref{TabSingle}, which achieved R$^2$ above 0.997 across all pollutants, both ML and DL approaches lagged behind. This underscores the advantage of GNNs in modeling spatial dependencies among monitoring stations, an aspect that standard ML and DL models do not explicitly leverage.
}

```latex
\begin{table}[h!]
\centering
\small
\caption{Performance of ML and DL models for predicting PM$_{1}$, PM$_{10}$, and PM$_{2.5}$.}
\label{tab:ml-dl-performance}
\begin{tabular}{llrrr}
\toprule
Pollutant & Model & MAE & RMSE & R$^{2}$ \\
\midrule
PM$_{2.5}$ & kNN     & 2.366 & 17.391 & 0.699 \\
           & RF      & 1.363 & 6.607  & 0.886 \\
           & ET      & 1.347 & 6.252  & 0.892 \\
           & DT      & 1.902 & 13.449 & 0.767 \\
           & GB      & 1.326 & 6.259  & 0.892 \\
           & Prophet & 2.800 & 4.950  & 0.669 \\
           & LSTM    & 2.450 & 23.690 & 0.680 \\
           & GRU     & 3.072 & 5.014  & 0.661 \\
\midrule
PM$_{10}$ & kNN     & 2.599 & 20.737 & 0.717 \\
          & RF      & 1.496 & 8.002  & 0.891 \\
          & ET      & 1.509 & 8.041  & 0.890 \\
          & DT      & 2.048 & 15.218 & 0.792 \\
          & GB      & 1.437 & 7.544  & 0.897 \\
          & Prophet & 3.140 & 5.590  & 0.673 \\
          & LSTM    & 2.830 & 31.250 & 0.673 \\
          & GRU     & 3.495 & 5.816  & 0.646 \\
\midrule
PM$_{1}$ & kNN     & 1.679 & 8.586  & 0.630 \\
         & RF      & 0.902 & 2.832  & 0.878 \\
         & ET      & 0.888 & 2.646  & 0.886 \\
         & DT      & 1.192 & 4.947  & 0.787 \\
         & GB      & 0.879 & 2.642  & 0.886 \\
         & Prophet & 1.840 & 3.200  & 0.683 \\
         & LSTM    & 1.550 & 10.020 & 0.689 \\
         & GRU     & 2.041 & 3.279  & 0.666 \\
\bottomrule
\end{tabular}
\end{table}
```


\medskip 
{\color{black}Figure~\ref{fig:avg-results-1} provides a comparative evaluation of the averaged prediction performance across GNN models, traditional ML models, and DL models (Prophet, LSTM, GRU) for PM concentration prediction.

\begin{figure}[h!]
    \centering
    \includegraphics[width=15cm]{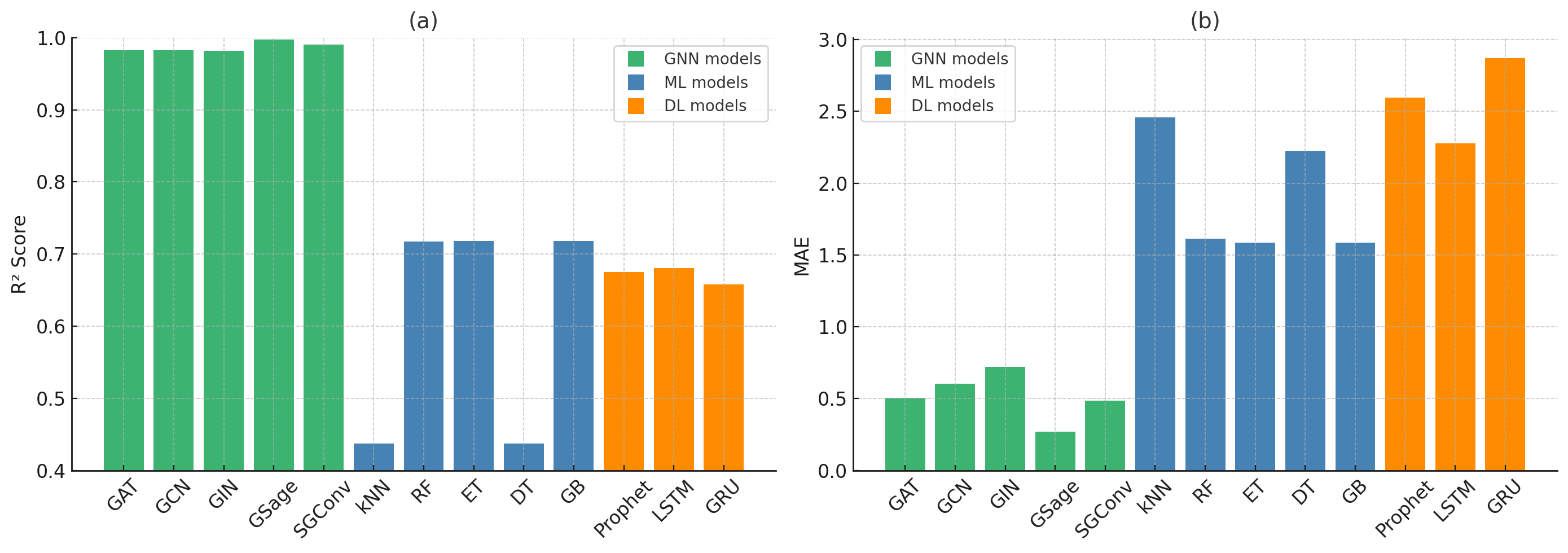}
    \caption{\color{black} Averaged prediction performance (R$^{2}$ and MAE) comparison among GNN models, traditional ML models, and DL models for PM concentration prediction. }
    \label{fig:avg-results-1}
\end{figure}

Results indicate that GNN models significantly outperform both ML and DL models. GSage demonstrates the highest predictive accuracy with an $R^{2}$ of 0.9974 and the lowest MAE of 0.2684. SGConv and GAT also show strong performance, with $R^{2}$ scores exceeding 0.98. In comparison, traditional ML models such as kNN and DT yield notably lower $R^{2}$ values (0.4370) and higher MAEs (2.459 and 2.222, respectively). Among DL models, LSTM achieves better performance than GRU and Prophet, with an average $R^{2}$ of 0.6807 and MAE of 2.277. However, these results remain inferior to those of GNN-based models, demonstrating the effectiveness of graph-based learning in modeling spatial and temporal dependencies in air pollution prediction.}

\medskip
The superior performance of GNN models over traditional machine learning and deep learning methods can be attributed to their ability to explicitly model spatial dependencies among air quality monitoring stations. While conventional ML models such as kNN and decision trees treat each observation independently, failing to account for spatial interactions, and DL models like LSTM and GRU primarily focus on capturing temporal patterns, GNNs leverage graph structures to integrate both spatial and contextual relationships. Specifically, GSage aggregates information from neighboring nodes through adaptive message passing, enabling it to learn richer and more context-aware spatial representations. In contrast, models like kNN, DT, and Prophet operate in isolation and lack mechanisms to incorporate spatially structured information, resulting in lower prediction accuracy. This spatial-awareness, enabled through automatically constructed graphs derived from confusion matrices, empowers GNNs to model the complex interplay between monitoring stations and leads to their consistently superior predictive performance.

\medskip 
\subsubsection{Multi-Step Prediction of PM Levels Using GNNs}
The performance of the investigated GNN models for the multi-step prediction of PM concentrations (PM${1}$, PM${2.5}$, and PM$_{10}$) was assessed. The GNN-based forecasting approach in this study encompasses multiple horizons, ranging from 3 to 12 hours ahead, with predictions made on an hourly basis. To capture trends at different time granularities, data is aggregated into hourly bins of varying lengths: 3, 6, 9, and 12 hours. Table~\ref{tab:multiple-forecast} presents the results based on the testing data, evaluating each GNN model (SimpleGCN, GCN, GIN, GAT, and GraphSage) for short- to medium-term PM concentration forecasting (3, 6, 9, and 12 hours ahead).

\begin{table}[h!]
\centering
\footnotesize
\caption{Performance of GNN models in predicting pollutant concentrations (PM$_{1}$, PM$_{2.5}$, and PM$_{10}$) for 3-hour, 6-hour, 9-hour, and 12-hour ahead predictions.} \label{tab:multiple-forecast}
\begin{tabular}{ll|cc|cc}
\toprule
Pollutant & Model & \multicolumn{2}{c|}{3H - 6H} & \multicolumn{2}{c}{9H - 12H} \\
          &       & $R^{2}$ & $R^{2}$ & $R^{2}$ & $R^{2}$ \\\toprule

\multirow{5}{*}{PM$_{1}$}  
 & GCN   & 0.9796 & 0.9349 & 0.9431 & 0.9139 \\
 & GAT   & 0.9872 & 0.9511 & 0.9600 & 0.9583 \\
 & GIN   & 0.9784 & 0.9686 & 0.9495 & 0.9432 \\
 & GSage & 0.9919 & 0.9822 & 0.9717 & 0.9779 \\
 & SGConv & 0.9851 & 0.9584 & 0.9442 & 0.9409 \\\midrule

\multirow{5}{*}{PM$_{10}$}  
 & GCN   & 0.9853 & 0.9499 & 0.9513 & 0.9296 \\
 & GAT   & 0.9905 & 0.9745 & 0.9767 & 0.9386 \\
 & GIN   & 0.9875 & 0.9665 & 0.9507 & 0.9523 \\
 & GSage & 0.9959 & 0.9854 & 0.9573 & 0.9804 \\
 & SGConv & 0.9867 & 0.9697 & 0.9528 & 0.9449 \\\midrule

\multirow{5}{*}{PM$_{2.5}$}  
 & GCN   & 0.9828 & 0.9498 & 0.9568 & 0.9375 \\
 & GAT   & 0.9889 & 0.9650 & 0.9722 & 0.9646 \\
 & GIN   & 0.9851 & 0.9786 & 0.9481 & 0.9118 \\
 & GSage & 0.9937 & 0.9851 & 0.9756 & 0.9799 \\
 & SGConv & 0.9893 & 0.9579 & 0.9483 & 0.9395 \\\toprule
\end{tabular}
\end{table}

\medskip 
The results in Table~\ref{tab:multiple-forecast} show that all GNN models perform well in predicting PM concentrations, with the $R^{2}$ scores indicating high predictive accuracy across different time horizons. For short-term predictions (3 and 6 hours ahead), the GSage model consistently achieves the highest $R^{2}$ scores across all pollutants, demonstrating superior performance. In medium-term predictions (9 and 12 hours ahead), the GSage model also shows strong performance, but other models like GAT and GCN exhibit competitive results. Across both experimental series, GraphSage emerged as the top performer for PM prediction. GraphSage excels due to its efficiency, flexibility, and generalizability. Unlike GNNs limited to static graphs, GraphSage supports inductive learning, allowing predictions on new nodes and graphs, ideal for dynamic datasets. Its neighborhood sampling technique reduces computational complexity and memory requirements, making it efficient for massive graphs. GraphSage's customizable message-passing scheme allows for defining specific aggregation functions suitable for various tasks such as link prediction, graph clustering, and anomaly detection. It also provides interpretability through learnable aggregation functions, offering insights into how the model uses information from neighboring nodes. GraphSage's efficient message-passing and sampling techniques also result in faster training times. In contrast, GATs focus on the relative importance of neighboring nodes, providing a powerful approach for tasks where this is critical. While offering interpretability through attention mechanisms, GATs require more computational resources than GraphSage.

\medskip
To comprehensively assess model performance, this section also includes a comparative analysis between GNN-based, DL-based, and traditional ML-based approaches for multi-step PM forecasting. By evaluating all methodologies under the same predictive settings, we aim to highlight their respective strengths and limitations in handling the spatial and temporal complexities of air pollution data. Table~\ref{tab} shows the performance of ML and DL models in multi-step PM forecasting across different prediction horizons (3, 6, 9, and 12 hours). The DL models include Prophet, LSTM, and GRU, which are widely used for temporal sequence modeling and provide a useful benchmark for evaluating the added value of graph-based learning.

```latex
\begin{table}[h!]
\centering
\footnotesize
\caption{Performance of ML and DL models in predicting PM$_{1}$, PM$_{2.5}$, and PM$_{10}$ concentrations for 3-hour, 6-hour, 9-hour, and 12-hour forecasts.}
\label{tab:ml-dl-multistep}
\begin{tabular}{ll|cc|cc}
\toprule
Pollutant & Model & \multicolumn{2}{c|}{3H - 6H} & \multicolumn{2}{c}{9H - 12H} \\
          &       & $R^{2}$ & $R^{2}$ & $R^{2}$ & $R^{2}$ \\
\toprule

\multirow{8}{*}{PM$_{2.5}$}
 & kNN     & 0.699 & 0.624 & 0.439 & 0.588 \\
 & RF      & 0.886 & 0.831 & 0.705 & 0.819 \\
 & ET      & 0.892 & 0.832 & 0.718 & 0.805 \\
 & DT      & 0.767 & 0.682 & 0.485 & 0.600 \\
 & GB      & 0.892 & 0.831 & 0.725 & 0.819 \\
 & Prophet & 0.796 & 0.736 & 0.681 & 0.633 \\
 & LSTM    & 0.813 & 0.692 & 0.688 & 0.636 \\
 & GRU     & 0.776 & 0.618 & 0.666 & 0.617 \\
\midrule

\multirow{8}{*}{PM$_{10}$}
 & kNN     & 0.717 & 0.621 & 0.527 & 0.588 \\
 & RF      & 0.891 & 0.837 & 0.724 & 0.810 \\
 & ET      & 0.890 & 0.837 & 0.743 & 0.817 \\
 & DT      & 0.792 & 0.758 & 0.360 & 0.593 \\
 & GB      & 0.897 & 0.845 & 0.736 & 0.814 \\
 & Prophet & 0.802 & 0.739 & 0.684 & 0.633 \\
 & LSTM    & 0.812 & 0.618 & 0.684 & 0.640 \\
 & GRU     & 0.801 & 0.664 & 0.654 & 0.422 \\
\midrule

\multirow{8}{*}{PM$_{1}$}
 & kNN     & 0.630 & 0.565 & 0.361 & 0.532 \\
 & RF      & 0.878 & 0.821 & 0.676 & 0.800 \\
 & ET      & 0.886 & 0.822 & 0.691 & 0.798 \\
 & DT      & 0.787 & 0.737 & 0.466 & 0.602 \\
 & GB      & 0.886 & 0.830 & 0.694 & 0.795 \\
 & Prophet & 0.787 & 0.732 & 0.678 & 0.634 \\
 & LSTM    & 0.801 & 0.609 & 0.687 & 0.634 \\
 & GRU     & 0.802 & 0.660 & 0.561 & 0.620 \\
\toprule
\end{tabular}
\end{table}
```

\medskip
The results in Table~\ref{tab:ml-dl-multistep} reveal that ML models generally exhibit lower $R^{2}$ values across all pollutants and prediction horizons compared to GNNs. For short-term predictions (3-hour and 6-hour ahead), the best-performing ML models (RF, ET, and GB) achieve $R^{2}$ values ranging from 0.705 to 0.892, which is notably lower than GNN models such as GSage and GAT, which exceed 0.99. DT and kNN consistently underperform, particularly for PM${1}$ and PM${2.5}$, with $R^{2}$ values below 0.7, indicating a weaker ability to capture spatiotemporal dependencies.  Incorporating DL models, Prophet, LSTM, and GRU, into the evaluation provides further insight into their predictive capabilities. While Prophet achieves moderate short-term accuracy (e.g., $R^2$ around 0.79–0.80), its performance declines over longer horizons. LSTM and GRU models show stronger results than traditional ML methods in some cases, particularly at the 3-hour mark (with $R^2$ above 0.80), but their accuracy degrades significantly as the prediction window extends, with $R^2$ dropping below 0.65 in most 12-hour scenarios. For longer prediction horizons (9-hour and 12-hour ahead), the gap between ML and DL models and GNN models further widens. GNNs maintain strong predictive accuracy, while ML and DL models show a more significant decline in $R^{2}$, particularly for PM$_{10}$. For instance, the best non-GNN models (RF and GB) achieve an $R^{2}$ of 0.819–0.825, and LSTM and GRU fall below 0.65, whereas GraphSage maintains values above 0.97, showing superior generalization over extended forecasting windows.

\begin{figure}[h!]
    \centering
    \includegraphics[width=14cm]{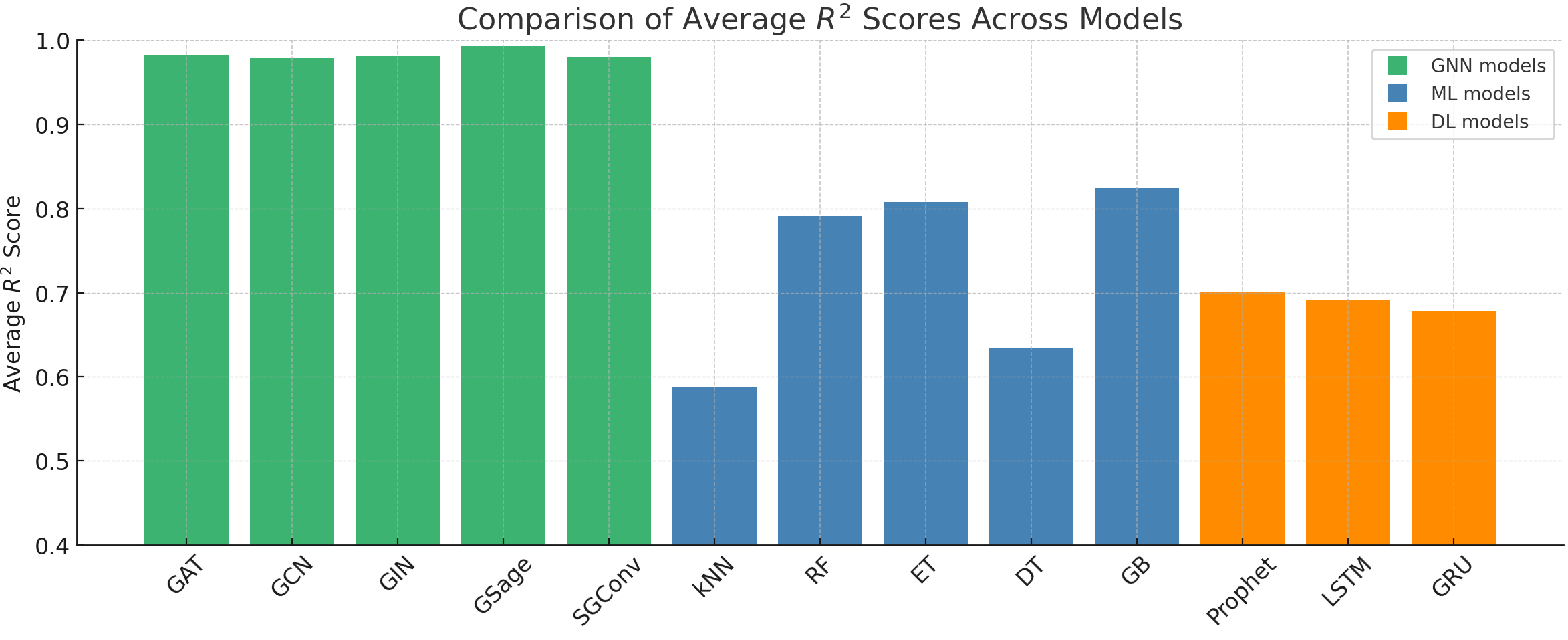}
    \vspace{-0.3cm}
    \caption{Comparison of the averaged R$^{2}$ scores for different models in predicting PM concentrations. Green bars represent GNN models, blue bars denote traditional machine learning models, and orange bars correspond to deep learning (DL) models. Higher R$^{2}$ values indicate better predictive performance.}\label{fig:R2comparison}
\end{figure}

\medskip  
Figure~\ref{fig:R2comparison} presents a comparative analysis of the predictive performance of the considered models using the \( R^2 \) metric. The results demonstrate that GNN-based models consistently outperform traditional ML models in predicting particulate matter concentrations. Among the GNNs, GSage achieves the highest \( R^2 \) score (0.9814), indicating its superior ability to capture spatial dependencies within the air pollution dataset. The strong performance of GSage can be attributed to its inductive learning capability and neighborhood sampling approach, which enhances generalization and computational efficiency. Other GNN models, including GAT (0.9690), GIN (0.9600), SGConv (0.9598), and GCN (0.9512), also exhibit high predictive accuracy, reinforcing the effectiveness of graph-based learning in air quality forecasting. These models effectively capture complex spatial correlations among monitoring stations, leading to more robust predictions. Conversely, traditional ML models show lower \( R^2 \) scores, with kNN (0.5742) and DT (0.6358) performing the worst. The relatively poor performance of these models highlights their limitations in handling spatial dependencies and complex relationships within the dataset. Among ML approaches, ET (0.8110), RF (0.8063), and GB (0.8137) achieve moderate \( R^2 \) scores, benefiting from ensemble learning techniques that improve stability and predictive power. In addition, deep learning models were included to provide a broader benchmark. Prophet achieved an average \( R^2 \) score of 0.7131, followed by GRU (0.6761) and LSTM (0.6828). Although these models capture temporal dependencies effectively, their inability to model spatial interactions limits their performance compared to GNNs.

\begin{figure}[h!]
    \centering
    \includegraphics[width=0.475\linewidth]{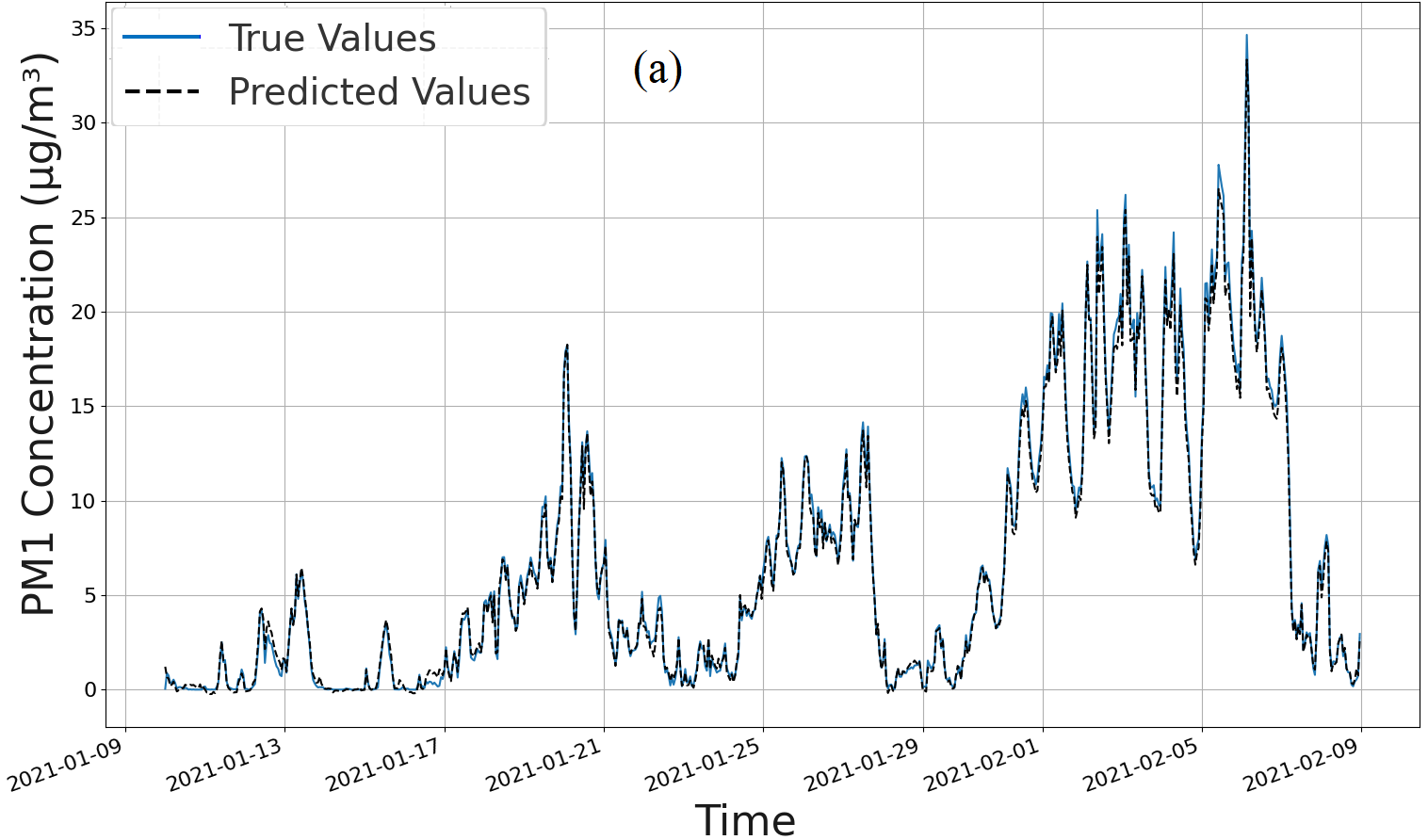}
     \includegraphics[width=0.475\linewidth]{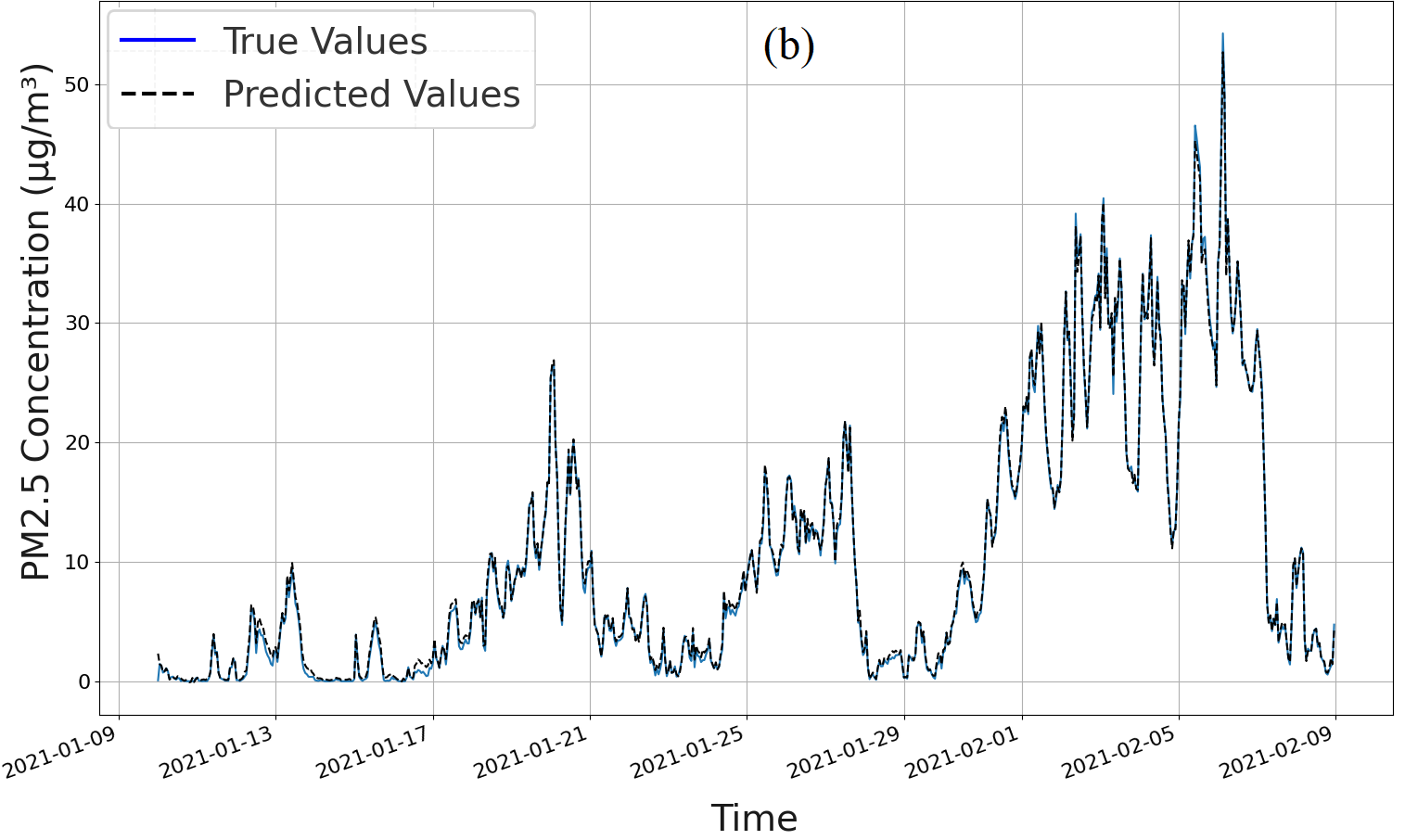}
    \includegraphics[width=0.475\linewidth]{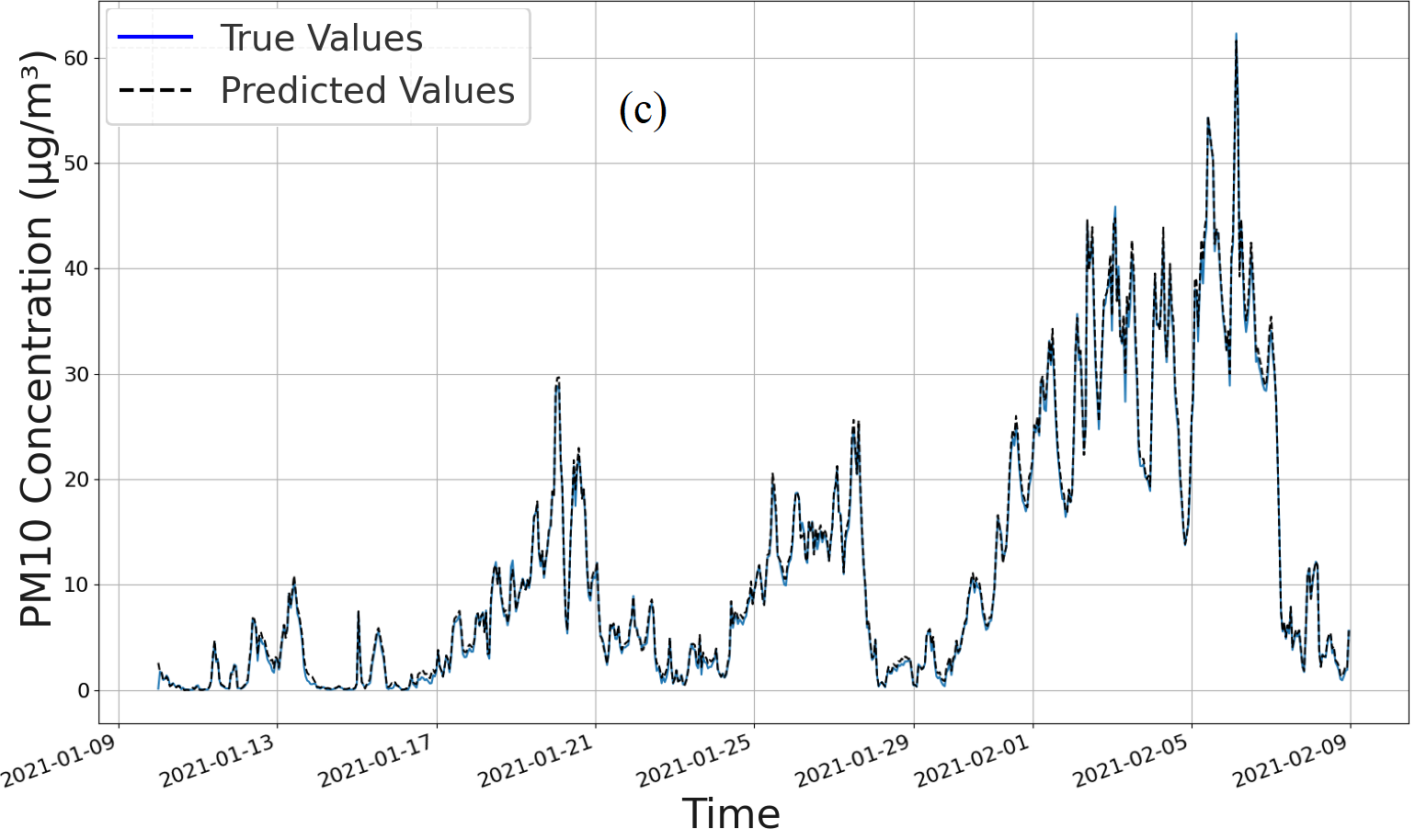}
    \caption{Comparison of predicted and observed pollutant concentrations using the GraphSage model for (a) PM${1}$, (b) PM${2.5}$, and (c) PM$_{10}$. The predictions closely follow the observed trends, demonstrating the model’s ability to capture temporal variations in pollutant concentrations.}
    \label{fig:PM-GSage}
\end{figure}

\medskip 
Figure~\ref{fig:PM-GSage} (a-c) presents a comparison between the predicted and observed pollutant concentrations for PM${1}$, PM${2.5}$, and PM$_{10}$, respectively, using the best-performing model, GraphSage. The predicted values closely follow the observed trends, demonstrating the model's ability to capture temporal variations and fluctuations in pollutant concentrations. The consistency between the two curves indicates that GraphSage effectively learns spatial and temporal dependencies, leading to accurate short-term forecasts. These results highlight the effectiveness of GNN-based approaches in air pollution forecasting.

\medskip
To evaluate computational efficiency for real-time forecasting, we measured the average execution time required for making predictions with each trained GNN model. Since training is performed offline, the reported values reflect only the online inference time per sample. The results show that the proposed models are computationally lightweight. Specifically, GraphSage achieved the fastest execution time of 0.0017 seconds, followed by GIN (0.0018 s), SGConv (0.0053 s), and GCN (0.0075 s). GAT, due to its attention mechanism, recorded a slightly higher execution time of 0.0234 seconds. These fast inference times are partially attributed to the relatively small graph size used in this study, with only 13 monitoring stations, and to the fact that predictions are made at an hourly resolution. The proposed approach is not designed for real-time streaming but for short-term hourly prediction, which further justifies its low computational requirements. Overall, the GNN models—especially GraphSage and GIN—offer a strong balance between accuracy and efficiency, making them suitable for near-real-time air quality forecasting applications.

\newpage
\subsection{Explainable GNN}
\begin{figure}[h!]
    \centering
    \includegraphics[width=8cm]{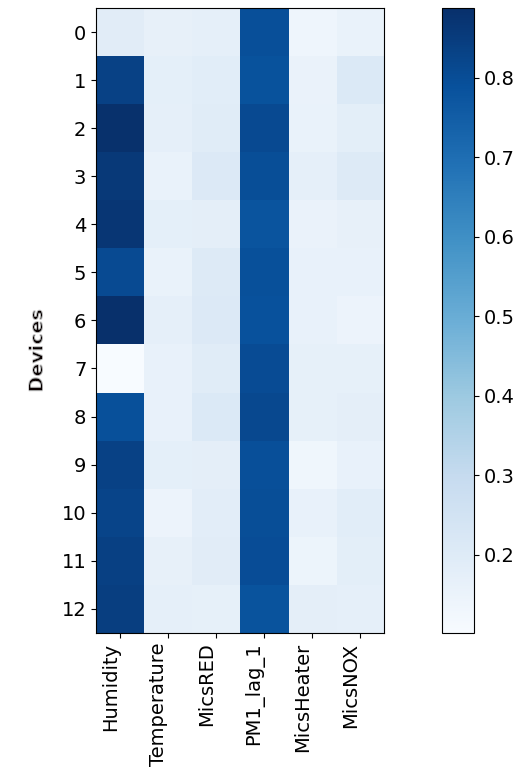}
    \caption{GSage features importance per device using GNNExplainer}
        \label{fig:GSage-features}
\end{figure}

This section presents the findings related to the explainability of the experimental results. Two GNN explanation methods were employed: the Graph Neural Network Explainer (GNNExplainer) for analyzing feature importance~\cite{ying2019gnnexplainer} and the Parameterized Explainer for GNNs (PGExplainer) for analyzing the importance of graph structure~\cite{luo2020parameterized}, specifically the edges. The GNNExplainer method was used to identify the most influential features for predicting PM$_{1}$ concentration. As illustrated in Figure~\ref{fig:GSage-features}, the lagged PM$_{1}$ feature and humidity emerged as the most influential features of the GraphSage model. During the graph regression task, these features were consistently important across all monitoring stations (nodes). This insight highlights the significance of temporal dependencies and environmental factors in the model's predictive performance.

\medskip 
PGExplainer, on the other hand, was utilized to analyze the importance of edges within the graph structure. By examining the connections between nodes, PGExplainer provided insights into how the relationships between different monitoring stations influence the model's predictions. Understanding edge importance helps in interpreting the model's reliance on specific inter-device relationships, further elucidating the underlying dynamics captured by the GNN. 

\begin{figure}[h!]
        \centering
        \includegraphics[width=8.5cm]{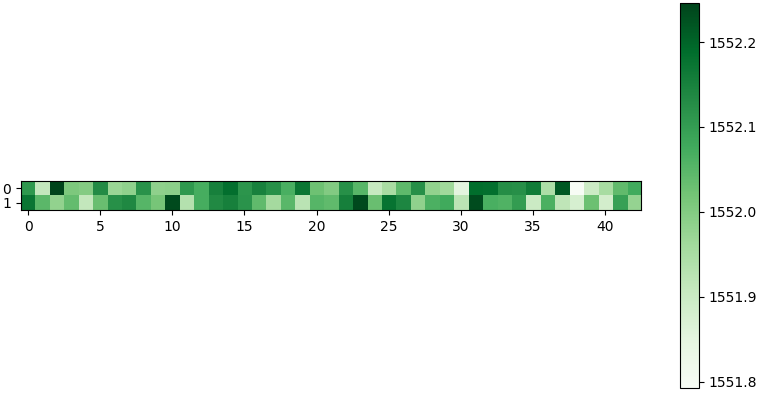}
        \caption{Explaining Edges importance using PGExplainer}
        \label{fig:explain-GSage-edges}
    \end{figure}

\medskip 
Figure~\ref{fig:explain-GSage-edges} visualizes the importance of edges in our graph using a heatmap derived from the PGExplainer explanation. Darker regions represent edges that significantly influence the model's predictions. The first row depicts the influence of edges in one direction, while the second row shows the influence in the opposite direction. This visualization clearly demonstrates the directional significance of certain edges in the graph, providing a deeper understanding of how the structure and connectivity within the graph contribute to the model's predictive performance.

These explainability methods enhance the transparency of GNN models by shedding light on the critical features and structural elements that drive their predictions. This improved understanding can inform model refinement and deployment strategies, ensuring more reliable and interpretable forecasting outcomes.

\begin{figure} [h!]
    \centering
        \includegraphics[width=7cm]{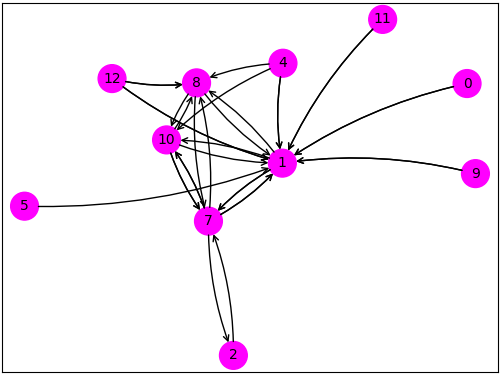}
        \includegraphics[width=7cm]{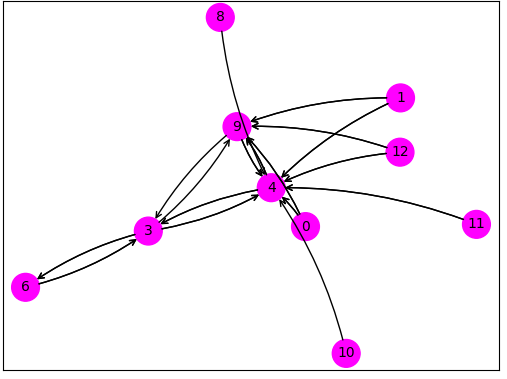}
\caption{Example of two sub-graphs used to highlight the neighboring importance of nodes 2 and 6, respectively}
\label{fig:sub_graph_x}
\end{figure}

\medskip 
Figure~\ref{fig:sub_graph_x}  presents subgraphs extracted by GNNExplainer, emphasizing the importance of neighboring nodes for monitoring stations 2 and 6. In Figure~\ref{fig:graph}, which shows the original graph, there is a bidirectional connection between monitoring stations 6 and 3. However, the GNNExplainer analysis reveals that device 3 relied on a reduced set of neighboring nodes (see Figure~\ref{fig:sub_graph_x} (right)) compared to the original graph for the device 6 regression task. This reduced set was still relevant for the prediction, according to the explainer, indicating that not all original connections were necessary for accurate forecasting.

A similar behavior is observed between monitoring stations 2 (Figure~\ref{fig:sub_graph_x}, left panel) and 7. While Figure~\ref{fig:graph} shows their connection in the original graph, the GNNExplainer analysis (Figure~\ref{fig:sub_graph_x}, left panel) reveals that device 7 relied on a more compact set of neighboring nodes for the regression task. This suggests that a full connection wasn’t necessary for accurate prediction, and the model effectively identified the most critical relationships within the graph.

These findings underscore the effectiveness of GNNExplainer in highlighting the essential nodes and connections that contribute most significantly to the model's predictions. By focusing on the most influential neighbors, the explainer provides insights into the simplified yet impactful structure used by the GNN for its regression tasks, enhancing our understanding of the model's decision-making process.

\section{Conclusion}\label{sec5}
This paper presents an advanced approach to predict particulate matter (PM) concentrations across multiple monitoring sites using GNNs. The proposed method integrates automatic graph construction, hybrid loss functions, and explainability techniques to enhance both accuracy and interpretability in PM forecasting. An efficient graph construction method using a confusion matrix from supervised learning enables the model to automatically construct graphs and enhance predictive performance. Additionally, a hybrid loss function combining energy distance and Huber loss mitigates vanishing gradient issues, ensuring stable and efficient learning. Extensive experiments were conducted using real-world datasets to validate the effectiveness of the approach, comparing five GNN architectures: GCN, GAT, GIN, SGConv, and GraphSage. Both single-step and multi-step forecasting results demonstrated that GraphSage achieved the highest accuracy, followed by GAT and SGConv. To ensure transparency, GNNExplainer and PGExplainer were utilized to analyze feature importance and graph structures.   Beyond evaluating GNN models, a comparative analysis with traditional ML models, including kNN, RF, ET, DT, and GB, was performed. The results highlighted the superior performance of GNN-based approaches, particularly in capturing spatial dependencies and adapting to dynamic patterns in air pollution data. While some ML models, such as Gradient Boosting and Extra Trees, showed competitive performance in short-term predictions, they struggled to maintain accuracy over extended forecasting horizons. {\color{black}In addition, this study included a comparison with three deep learning models: Prophet, LSTM, and GRU. While these models demonstrated reasonable accuracy in short-term forecasting scenarios, they exhibited higher error rates and reduced $R^2$ scores compared to GNNs, particularly for longer forecasting horizons. }

\medskip
{\color{black} While this study focused on data from a single urban area (Salt Lake City), extending the proposed framework to other geographical regions with different pollution dynamics represents an important direction for future research. The data-driven nature of the approach, particularly the automatic graph construction based on confusion matrices, enables it to adapt to diverse spatial and temporal patterns without relying on predefined structures. Future work will evaluate the framework across multiple cities and pollution contexts to assess its generalizability and scalability. Furthermore,  although this study focuses on spatial and short-to-medium-term temporal dependencies using historical PM measurements, future work can enhance the proposed model by incorporating additional environmental variables such as wind, rainfall, temperature, and urban morphology. These factors can be encoded as node or edge attributes to better capture the complex dynamics of pollution diffusion. Additionally, extending the temporal scope of the analysis to cover seasonal and yearly trends would provide deeper insights into long-term air quality patterns and improve the robustness of the forecasting framework.

}

\bibliographystyle{unsrt}  
 \bibliography{references}  

\end{document}